\documentclass[letter,11pt]{article}

\usepackage[utf8]{inputenc}
\usepackage[T1]{fontenc}

\usepackage[a4paper,margin=1in]{geometry}

\usepackage{color}
\usepackage[colorlinks=true,linkcolor=blue,citecolor=blue]{hyperref}
\usepackage{amsmath, amssymb, amsthm}
\usepackage{mathtools}
\usepackage{graphics}
\usepackage{pifont}
\usepackage{tikz}
\usepackage{bbm, bm}
\usepackage{lipsum}

\usetikzlibrary{arrows.meta}
\usepackage{environ}
\usepackage{framed}
\usepackage{url}
\usepackage[framemethod=tikz]{mdframed}
\usepackage{appendix}
\usepackage{graphicx}
\usepackage{tcolorbox}
\allowdisplaybreaks[1]
\usepackage{nicefrac}
\usepackage{natbib}
\usepackage{thm-restate}
\usepackage[noabbrev,capitalize,nameinlink]{cleveref}
\crefname{equation}{}{}

\newcommand\remove[1]{}

\newtheorem{lemma}{Lemma}[section]
\newtheorem*{lemma*}{Lemma}
\newtheorem{theorem}[lemma]{Theorem}

\newtheorem*{corollary*}{Corollary}

\newtheorem{proposition}{Proposition}
\newtheorem{remark}{Remark}
\newtheorem{example}[lemma]{Example}

\newtheorem*{theorem*}{Theorem}
\newtheorem*{inducthyp*}{Inductive Hypothesis}
\newtheorem*{definition*}{Definition}
\newtheorem{definition}[lemma]{Definition}

\newtheorem*{rem*}{Remark}

\newcommand\N{\mathbb{N}}

\newcommand\E{\mathbb{E}}
\newcommand{\bitset}{\{0,1\}}

\newcommand{\eps}{\varepsilon}

\renewcommand{\forall}{\mathrm{\text{ for all }}}

\newcommand{\D}{\mathcal{D}}

\renewcommand{\H}{\mathcal{H}}

\newcommand{\A}{\mathcal{A}}

\renewcommand{\hat}{\widehat}

\DeclareFontFamily{U}{mathb}{\hyphenchar\font45}
\DeclareFontShape{U}{mathb}{m}{n}{<5> <6> <7> <8> <9> <10> gen * mathb
<10.95> mathb10 <12> <14.4> <17.28> <20.74> <24.88> mathb12}{}
\DeclareSymbolFont{mathb}{U}{mathb}{m}{n}
\DeclareMathSymbol{\rcirclearrow}{\mathbin}{mathb}{'367}

\foreach \x in {a,...,z}{%
	\expandafter\xdef\csname b\x\endcsname{\noexpand\mathbf{\x}}
}

\foreach \x in {A,...,Z}{%
	\expandafter\xdef\csname m\x\endcsname{\noexpand\mathbf{\x}}
}

\foreach \x in {A,...,Z}{%
	\expandafter\xdef\csname c\x\endcsname{\noexpand\mathcal{\x}}
}

\newif\ifrandom
\randomtrue

\newcommand{\KL}{{\mathrm{D_{KL}}}}

\newcommand{\V}{\mathcal{V}}
\newcommand{\U}{\mathcal{U}}

\newcommand{\todolater}[1]{}

\DeclareUnicodeCharacter{2113}{$\ell$}

\usepackage{multirow}

\newcommand{\lr}[1]{\mathopen{}\left(#1\right)}

\newcommand{\lrbra}[1]{\mathopen{}\left[#1\right]}

\newcommand{\lrset}[1]{\mathopen{}\left\{#1\right\}}

\newcommand{\lrabs}[1]{\mathopen{}\left|#1\right|}

\newcommand{\VC}{\mathrm{VC}}

\newcommand{\1}{\mathbbm{1}}

\newcommand{\X}{\mathcal{X}}
\usepackage{algorithm}
\usepackage{algpseudocode}

\newcommand{\er}{\mathrm{er}}
\newcommand{\uVCE}{\mathrm{uVCE}}
\newcommand{\VCE}{\mathrm{VCE}}
\newcommand{\DIS}{\mathrm{DIS}}
\newcommand{\SH}{\mathrm{SH}}
\newcommand{\Y}{\mathcal{Y}}
\renewcommand{\P}{\mathbb{P}}

\newcommand{\qconst}[1]{{\color{red!75!black}#1}}    
\newcommand{\qlabel}[1]{{\color{green!45!black}#1}}  
\newcommand{\qmarg}[1]{{\color{blue}#1}}             

\usepackage{amsmath}
\DeclareMathAlphabet{\mathbfcal}{OMS}{cmsy}{b}{n}

\title{Agnostic Marginal-Nonuniform Learning}

\author{
    Tanmay Devale \thanks{Department of Computer Science, Purdue University; \texttt{tdevale@purdue.edu}} 
    \and  Steve Hanneke \thanks{Department of Computer Science, Purdue University; \texttt{steve.hanneke@gmail.com}}
    \and  Amirreza Shaeiri \thanks{Department of Computer Science, Purdue University; \texttt{amirreza.shaeiri@gmail.com}}
}
\usepackage{times}

\begin{document}

\maketitle

\begin{abstract}%
  We study the problem of binary classification in the marginal-nonuniform learning framework in the agnostic setting.
  In this model, the constants in the learning rate may depend on the marginal distribution of the unlabeled data, but the guarantee must hold uniformly over all conditional distributions of the labels. We also study fine-grained rates.
  
  We consider two criteria.
  (1) The standard agnostic setting, where the learner competes with the best concept in the class. In this setting, an immediate consequence of standard agnostic PAC learning results gives us the optimal rate of $\tfrac{1}{\sqrt{n}}$ for every class $\H$ with finite $\VC(\H)$ and shows that arbitrarily slow rates are necessary otherwise. This also gives us the optimal fine-grained rate of $\sqrt{\tfrac{\VC(\H)}{n}}$. Therefore, the dependence on the marginal distribution does not help, which is in contrast to the realizable setting.
  (2) In the relaxed agnostic setting, where the learner competes with $(1+\lambda)$ times the error of the best concept for a fixed $\lambda \in (0,1]$, the picture from the realizable setting reappears. We obtain a trichotomy of the achievable optimal rates: exponential rates of the form $e^{-n}$ arise precisely when the class is finite; linear rates of the form $\tfrac{1}{n}$ arise when the class is infinite but $\VC(\H)$ is finite; and arbitrarily slow rates occur when $\VC(\H)=\infty$. We also obtain the optimal fine-grained rate of $\tfrac{\uVCE(\H)}{\lambda n}$, where $\uVCE(\H)$ is a recently introduced combinatorial complexity dimension called the uncentered $\VC$-eluder dimension.~\footnote{All results and proofs in this paper were obtained by the authors. AI assistance was limited to supporting the writing process.}\\
\end{abstract}


\section{Introduction}

A central question in statistical learning theory is how many labeled examples are needed to predict well on unseen data. 
The classical PAC model \citep{vc1971,valiant1984} answers the \emph{uniform} case: one rate has to hold for every marginal distribution over the instance space and every target concept in the class. 
This gives a clean worst-case theory, in which the VC dimension characterizes learnability, but it cannot tell an easy data distribution from a hard one. 
\emph{Marginal-nonuniform} learning \citep{bendavid1995,hanneke2025marginalnonuniform} relaxes exactly this point. 
The constants in the rate may depend on the marginal distribution $\D_\X$ of the unlabeled data, but once $\D_\X$ is fixed, the guarantee must hold uniformly over everything else. 
It sits between uniform PAC learning and universal learning \citep{bousquet2020theoryuniversallearning}, where the constants may depend on the whole data distribution.

Marginal-nonuniform learning has a classical counterpart in \emph{concept-nonuniform} learning, introduced by \citet{benedek1994} and usually called simply nonuniform learning. 
In this framework, the roles are exchanged: the constants in the rate may depend on the target concept, but the guarantee must hold uniformly over all marginal distributions. 
This model makes a much larger family of classes learnable, namely all countable unions of VC classes, and it underlies structural risk minimization, Occam's razor and the minimum description length principle \citep{vapnik1998,blumer1987,rissanen1978,lugosi1996}. 
It distinguishes concepts that are easy to learn from concepts that are hard to learn, whereas marginal-nonuniform learning distinguishes easy \emph{marginals} from hard ones. 
Together with uniform and universal learning, these are the four natural notions of learnability obtained by letting the rate depend on the marginal, on the target, on both, or on neither \citep{bendavid1995}. 
We recall them, in the form needed for this paper, in Section \ref{sec:main-results}.

In the realizable case, the possible marginal-nonuniform rates are by now well understood.
\cite{hanneke2025marginalnonuniform} proved a trichotomy for binary classification: every class is learnable at an exponential rate, at a linear rate, or only at arbitrarily slow rates, and in the linear case the optimal rate is $d/n$, where $d$ is the VC-eluder dimension of \cite{hanneke2025universalratesempiricalrisk}. 
This dimension can be much smaller than the VC dimension. 
The work of \cite{brukhim2026marginal} extended the characterization to multiclass classification, using the uncentered DS-eluder dimension. 
All of these results assume that some concept in the class labels the data perfectly.

\paragraph{Agnostic marginal-nonuniform learning.}
Consider an email provider that stores a large archive of messages. 
The prediction task is chosen only later: filtering spam, detecting phishing, or flagging messages as urgent. The messages, and hence the marginal $\D_\X$, are fixed before the task is known, which is the situation marginal-nonuniform guarantees are designed for. 
But the labels will come from users, who click carelessly and who disagree with each other: the same newsletter is spam to one user and wanted by another. 
No concept in the class will label the data perfectly. 
What the provider can hope for is a filter that is nearly as accurate as the \emph{best} concept in the class, whatever the task and the behavior of the users turn out to be.

This is the step from realizable to \emph{agnostic} learning \citep{Haussler1992,kearns1994}. 
It is the standard second step in PAC learning: one first studies the realizable case, where the optimal rate is $\VC(\H)/n$, and then drops the assumption on the labels, where the optimal rate becomes $\sqrt{\VC(\H)/n}$. We take the same step in the marginal-nonuniform setting. 
The learner competes with $\inf_{h\in\H}\er_\D(h)$, the constants may depend on the marginal $\D_\X$, and the guarantee must hold for \emph{every} distribution $\D$ over $\X\times\Y$ with that marginal. 
In the example, the archive fixes $\D_\X$, while the task and the behavior of the annotators, i.e., the conditional distribution of the label given the message, are left completely free.

If $\VC(\H) < \infty$, the classical agnostic PAC upper bound applies directly: it holds for every distribution $\D$ over $\X\times\Y$, with constants that do not depend on $\D$, so in particular it holds for every distribution with a given marginal $\D_\X$. 
The classical agnostic PAC lower bound also carries over. 
Its hard distributions all have the same marginal, namely the uniform distribution on a set of $\VC(\H)$ points shattered by $\H$, and only the conditional distribution of the labels depends on the sample size and on the learner. 
It is therefore a lower bound in the marginal-nonuniform model as well. 
Combining the two, $\H$ is agnostically marginal-nonuniformly learnable at exact rate $\frac{1}{\sqrt n}$ whenever $\VC(\H) < \infty$.
It also provides a fine-grained rate of $\sqrt{\frac{\VC(\H)}{n}}$.
Therefore, we could say that the dependence on the marginal does not help. 

\paragraph{Relaxed agnostic marginal-nonuniform learning.}
The $\frac{1}{\sqrt{n}}$ rate is the price of competing with the best concept \emph{exactly}. 
In many applications this is more than one needs. 
Suppose that in the example above the best rule in the class disagrees with the user on $1\%$ of the messages. 
A guarantee of the form ``error at most $1.1\%$, plus a term of order $1/n$'' is then far more useful than ``error at most $1\%$, plus a term of order $1/\sqrt n$''. 
When the best concept is already good, a small multiplicative loss costs almost nothing, while the additive $1/\sqrt n$ term dominates the bound.

This motivates the \emph{relaxed} agnostic criterion: for a fixed $\lambda \in (0,1]$, the learner competes with
$(1+\lambda)\inf_{h\in\H}\er_\D(h)$. 
Guarantees of this multiplicative form are common. 
The weighted majority algorithm competes with a constant factor times the loss of the best expert \citep{littlestone1994}; efficient algorithms for learning halfspaces with adversarial label noise under suitable distributional assumptions typically guarantee error $(1+\lambda)\mathrm{OPT} + \eps$ or $C\cdot\mathrm{OPT} + \eps$ \citep{daniely2015,awasthi2018powerlocalizationefficientlylearning}; and bounds that improve when the best error is small are known as optimistic or first-order bounds \citep{srebro2010}. 
The relaxed criterion also interpolates between the two classical settings: when the data are realizable it is the realizable criterion, and as $\lambda \to 0$ it becomes the agnostic one.

In this case we show that under the relaxed criterion the realizable picture returns, both for the rates and for the fine-grained rates. 
The achievable rates form the same trichotomy as in the realizable case \citep{hanneke2025marginalnonuniform}: exponential for finite classes, linear for infinite classes of finite VC dimension, and arbitrarily slow otherwise. 
Note that even a finite class only has rate $\frac{1}{\sqrt n}$ under the standard criterion, so the multiplicative slack already matters there. 
The fine-grained rate is of order $\frac{d}{\lambda n}$, where $d$ is the uncentered VC-eluder dimension of the class, and the dependence on the marginal is confined to an exponentially small term. 
So the leading term is again governed by the uncentered VC-eluder dimension rather than by the VC dimension. 
The $\sqrt{\VC(\H)/n}$ barrier of the agnostic case is therefore caused entirely by insisting on the constant $1$ in front of the best error.

\subsection{Main results}\label{sec:main-results}

We first place our setting among the four learning paradigms mentioned above. 
A distribution $\D$ over $\X\times\Y$ consists of its marginal $\D_\X$ and its conditional distribution of the label given the instance, which we denote $\D_{\Y|\X}$. 
In the realizable case, $\D_{\Y|\X}$ is given by a target concept; in general it plays the role that the target concept plays there. For $\lambda \in [0,1]$ and a learner $\A$, write
\[
    \mathcal E_\lambda(\A,\D,n) = \E_{S\sim\D^n}\lrbra{\er_\D(\A_S)} - (1+\lambda)\inf_{h\in\H}\er_\D(h)
\]
for the excess error, where $\lambda = 0$ is the standard agnostic criterion and $\lambda > 0$ the relaxed one.
Adapting \citet{bousquet2020theoryuniversallearning} and \citet{hanneke2025marginalnonuniform}, the four paradigms differ only in the order of the quantifiers. Throughout, we extend every rate function
$R:\mathbb{N}\to[0,1]$ to positive real arguments
by setting $R(t)=R(\lceil t\rceil)$ for all $t>0$.

\begin{definition}\label{def:paradigms}
    Let $R:\N\to[0,1]$ be a rate function with $\lim_{n\to\infty}R(n) = 0$ and let $\lambda\in[0,1]$.
    \begin{itemize}
    \item  $\H$ is \emph{$\lambda$-agnostically uniformly learnable at rate $R$} if
    \[
        \exists\, \A \ \ \qconst{\exists\, C,c>0} \ \ \qlabel{\forall\, \D_{\Y|\X}} \ \ \qmarg{\forall\, \D_\X} \ \ \forall\, n\in\N \;\;
        \mathcal E_\lambda(\A,\D,n) \le C R(cn).
    \]
    \item $\H$ is \emph{$\lambda$-agnostically universally learnable at rate $R$} if
    \[
        \exists\, \A \ \ \qlabel{\forall\, \D_{\Y|\X}} \ \ \qmarg{\forall\, \D_\X} \ \ \qconst{\exists\, C,c>0} \ \ \forall\, n\in\N \;\;
        \mathcal E_\lambda(\A,\D,n) \le C R(cn).
    \]
    \item $\H$ is \emph{$\lambda$-agnostically concept-nonuniformly learnable at rate $R$} if
    \[
        \exists\, \A \ \ \qlabel{\forall\, \D_{\Y|\X}} \ \ \qconst{\exists\, C,c>0} \ \ \qmarg{\forall\, \D_\X} \ \ \forall\, n\in\N \;\;
        \mathcal E_\lambda(\A,\D,n) \le C R(cn).
    \]
    \item $\H$ is \emph{$\lambda$-agnostically marginal-nonuniformly learnable at rate $R$} if
    \[
        \exists\, \A \ \ \qmarg{\forall\, \D_\X} \ \ \qconst{\exists\, C,c>0} \ \ \qlabel{\forall\, \D_{\Y|\X}} \ \ \forall\, n\in\N \;\;
        \mathcal E_\lambda(\A,\D,n) \le C R(cn).
    \]
    Here $\D$ is the distribution over $\X\times\Y$ with marginal $\D_\X$ and conditional label distribution
    $\D_{\Y|\X}$.
    \end{itemize}
\end{definition}

Uniform learning is the classical agnostic PAC model: the constants depend on nothing. 
In universal learning they may depend on the whole distribution. 
The two nonuniform models sit in between and are incomparable: in the concept-nonuniform model the constants may depend on how the labels are generated but not on the marginal, and in the marginal-nonuniform model it is the other way round. 
This paper is about the last paradigm. 
It is the natural one when the unlabeled data are fixed in advance and the labeling task is not.
We state our results at two levels of detail. The first is the \emph{rate} in the sense of
Definition \ref{def:paradigms}, which describes the shape of the learning curve. Throughout we assume $|\H| \ge 3$, which
only excludes trivial classes \citep{hanneke2025marginalnonuniform}.
Our first result is regarding the rates for standard agnostic and relaxed agnostic settings. 
\begin{theorem}\label{thm:intro-rates}
    [Rates, informal]
    Let $\H$ be a concept class with $|\H| \ge 3$.
    \begin{itemize}
        \item \emph{Standard agnostic learning.}
        \begin{enumerate}
            \item $\H$ is $0$-agnostically marginal-nonuniformly learnable at exact rate $\frac{1}{\sqrt{n}}$ if and
            only if $\VC(\H)<\infty$.
            \item $\H$ requires arbitrarily slow rates to be $0$-agnostically marginal-nonuniformly learnable if and
            only if $\VC(\H)=\infty$.
        \end{enumerate}
        \item \emph{Relaxed agnostic learning.} For every fixed $0 < \lambda \le 1$:
        \begin{enumerate}
            \item $\H$ is $\lambda$-agnostically marginal-nonuniformly learnable at exact rate $e^{-n}$ if and only if
            $\H$ is finite.
            \item $\H$ is $\lambda$-agnostically marginal-nonuniformly learnable at exact rate $\frac{1}{n}$ if and only
            if $\H$ is infinite and $\VC(\H)<\infty$.
            \item $\H$ requires arbitrarily slow rates to be $\lambda$-agnostically marginal-nonuniformly learnable if
            and only if $\VC(\H)=\infty$.
        \end{enumerate}
    \end{itemize}
\end{theorem}

For the standard criterion, the rates are exactly those of uniform agnostic PAC learning. 
For the relaxed criterion, they are exactly those of realizable marginal-nonuniform learning \citep{hanneke2025marginalnonuniform}.

Rates alone say nothing about the constants. 
A bound $C R(cn)$ in which $C$ and $c$ may depend on the marginal does not tell us how the class and the marginal together determine the error. 
This is addressed by the fine-grained rates of \citet{bousquet2022finegraineddistributiondependentlearningcurves},
which separate the two dependences into different additive terms: one asks for a bound of the form
\[
    \mathcal E_\lambda(\A,\D,n) \le K\,R(n) + \delta(n),
\]
where $K$ is a universal constant, the leading term $R(n)$ depends only on the class (and on $\lambda$),
and all dependence on the marginal is confined to a lower-order term $\delta(n) = o(R(n))$. A matching
lower bound asks that a universal constant multiple of $R(n)$ cannot be avoided. The marginal then
determines only how long one has to wait before the class-dependent rate takes over, and not the rate
itself. The precise definitions are in Section \ref{sec:prelims}.

\begin{theorem}[Fine-grained rates, informal]\label{thm:intro-fine}
    \begin{enumerate}
        \item If $\VC(\H) < \infty$, then $\H$ is $0$-agnostically marginal-nonuniformly learnable at exact fine-grained rate $\sqrt{\frac{\VC(\H)}{n}}$.
        \item If $1 \le  \uVCE(\H) < \infty$, then for every $0 < \lambda \le 1$, $\H$ is $\lambda$-agnostically marginal-nonuniformly learnable at exact fine-grained rate $\frac{\uVCE(\H)}{\lambda n}$.
    \end{enumerate}
\end{theorem}

\section{Related Works}

\paragraph{Agnostic universal and nonuniform learning.}
The work of \citet{hanneke2026universalagnostic} extends the universal learning theory of \citet{bousquet2020theoryuniversallearning} to the agnostic setting, showing that the optimal universal rate is $e^{-n}$, $e^{-o(n)}$, $o(n^{-1/2})$, or arbitrarily slow, and \citet{hanneke2025ermagnostic} characterize the agnostic universal rates of ERMs. Since the constants there may depend on the entire data distribution, every finite class is learnable at rate $e^{-n}$ even under the standard criterion, whereas in the marginal-nonuniform framework two concepts already force rate $1/\sqrt{n}$ unless the criterion is relaxed. In classical comparator-dependent nonuniform learning
\footnote{Classical agnostic nonuniform learning allows
the guarantee to depend on the comparator
$h\in\mathcal H$. This differs from Definition~1,
where the constants may depend on $D_{Y\mid X}$
but must hold uniformly over all marginals.
Standard structural risk minimization guarantees
do not directly imply guarantees under Definition~1.},
the agnostic case is handled by structural risk minimization \citep{vapnik1998,lugosi1996,shalev2014understanding}, and \citet{hanneke2025nonuniformbandit} characterize nonuniform multiclass learnability in both the realizable and the agnostic settings, under both full-information and bandit feedback. To our knowledge, marginal-nonuniform learning has so far been studied only in the realizable case \citep{hanneke2025marginalnonuniform, brukhim2026marginal}.

\paragraph{Multiplicative slack in the uniform setting.}
In the uniform learning framework, it is well known that a multiplicative slack yields fast rates. Relative deviation and optimistic bounds \citep{vapnik1998,srebro2010} show that the error of ERM on $n$ examples, over a class of VC dimension $d$, is at most $(1+\lambda)$ times the best error in the class plus $O\big(d\log(n)/(\lambda n)\big)$, and the first-order learner of \citet{mathiasen2026optimalagnosticpacalgorithm} (see also \citet{hanneke2024revisitingagnosticpaclearning}) removes the logarithmic factor. Additionally, the work of \citet{asilis2025agnostic} gives a learner whose error is at most $c$ times the best error in the class, for a constant $c\le 2.1$, plus the optimal additive term. In statistics, bounds of this form, with a leading constant larger than one, are called nonexact oracle inequalities, and they can give fast rates where exact ones only give slow rates \citep{lecue2012general}. Our results for the relaxed criterion are the marginal-nonuniform analogue, with the VC dimension replaced by the uncentered VC-eluder dimension, which can be much smaller, and with exponential rates for finite classes; neither follows from the uniform bounds.

\paragraph{Learning with a known marginal.}
Marginal-nonuniform learning is related to learning under a fixed marginal that is known to the learner \citep{benedek1991learnability}. This is also the idealized limit of semi-supervised learning, where the value of unlabeled data has been studied extensively \citep{bendavid2008does,darnstadt2013unlabeled,gopfert2019when}. Our fine-grained lower bounds use a single marginal, so they hold even for learners that know $\mathcal{D}_{\mathcal{X}}$. In particular, under the standard criterion, knowing the marginal does not improve the fine-grained rate $\sqrt{\mathrm{VC}(\mathcal{H})/n}$.

\section{Preliminaries}
\label{sec:prelims}
    Let $\X$ denote a non-empty set which we call the \emph{instance space}, let $\Y=\bitset$ denote our \emph{label space},  and let $\H \subseteq \Y^\X$ be our \emph{concept class}. The measure-theoretical technicalities have been addressed in appendix \ref{appendix:measurability}. A data distribution $\D$ is a distribution over $\X \times \Y$. For every data distributions $\D$ we denote its marginal distribution over $\X$ by $\D_\X$. 
    A \emph{learning algorithm} $\A$ maps a dataset $S \in \bigcup_{n=1}^\infty (\X\times\Y)^n$ to a classifier $\A_S : \X \to \Y$. We allow learning algorithms to use internal randomness, independent of the data; then $\A_S$ is a random classifier, and $\er_\D(\A_S)$ and all expectations below are also taken over this randomness.

    The \emph{error rate} of a classifier $h:\X \to \Y$ with respect to a data distribution $\D$ is $\er_{\D}(h) = \P_{(x,y)\sim \D}[h(x) \neq y]$. 
    Similarly, given a dataset $S \in (\X \times \Y)^n$ the error rate $\A_S$ is defined as $\er_\D(\A_S) = \P_{(x,y)\sim \D}[\A_S(x)\neq y]$ and for a concept $h \in \H, \er_\D(h) = \P[h(x) \neq y]$. 
    The works of \cite{hanneke2025marginalnonuniform, brukhim2026marginal} focused on the \emph{realizable} case: namely those distributions satisfying $\inf_{h \in \H}\er_\D(h)=0$ and study the rate of convergence of the expected error rate $\E\lrbra{\er_\D(\A_S)}$ holding uniformly over all label conditionals with the given marginal $\D_\X$.
    In contrast, in the context of $\lambda$-\emph{agnostic learning} for a fixed $\lambda \in [0,1]$ our interest is in characterizing the optimal rates of convergence of the expected \emph{excess} error rate $\E\lrbra{\er_\D(\A_S)}-(1+\lambda)\inf_{h \in \H}\er_\D(h)$ holding uniformly over all concept in our concept class for the distribution $\D$ (without realizability restrictions). 
    The case $\lambda = 0$ recovers the \emph{standard agnostic} setting.
    For $\lambda>0,$ we call this the \emph{relaxed agnostic} setting.
    We extend the definitions of achievable rates from the works of \cite{hanneke2025marginalnonuniform, brukhim2026marginal} to apply in the context of $\lambda$-agnostic learning. Importantly, in the following definitions, note that the order of the quantifiers allows for the constants $C,c$ to be $\D_\X$-dependent (which is what distinguishes marginal-nonuniform rates from uniform rates). 

    Now, we present the definition of agnostic marginal-nonuniform learnability
    \begin{definition}
        Let $\H \subseteq \Y^\X$ be a concept class and let $\mathcal R:\N \to [0,1]$ be a \emph{rate function} such that $\lim_{n \to \infty}\mathcal R(n)=0$. We say
        \begin{enumerate}
            \item $\H$ is \emph{$\lambda$-agnostically marginal-nonuniformly learnable at rate $\mathcal R$} if there is a learning algorithm $\A$ such that for all marginal distributions $\D_\X$ there exists $C,c>0$ such that for all distributions $\D$ with marginal distribution $\D_\X, \E_{S \sim \D^n}[\er_\D(\A_S)]-(1+\lambda)\inf_{h \in \H}\er_\D(h) \le C\mathcal R(cn)$ for all $n \in \N$.
            \item $\H$ is \emph{not $\lambda$-agnostically marginal-nonuniformly learnable at rate faster than $\mathcal R$}, if for every learning algorithm $\A$, there exists a marginal distribution $\D_\X$ such that there exists $C,c>0$ there exists infinitely many $n \in \N$ for each of which there exists a distribution $\D_n$ with marginal $\D_\X$ such that $\E_{S \sim \D^n_n}[\er_{\D_n}(\A_S)]-(1+\lambda)\inf_{h \in \H}\er_{\D_n}(h)\ge C\mathcal R(cn)$.
            \item $\H$ is \emph{$\lambda$-agnostically marginal-nonuniformly learnable at exact rate $\mathcal R$} if $\H$ is $\lambda$-agnostically marginal-nonuniformly learnable at $\mathcal R$ and not faster than rate $\mathcal R$.
            \item $\H$ requires at least \emph{arbitrarily slow rates to be $\lambda$-agnostically marginal-nonuniformly learnable} if for every rate function $\mathcal R'$, the class $\H$ is not $\lambda$-agnostically marginal-nonuniformly learnable at rate faster than $\mathcal  R'$.
        \end{enumerate}
    \end{definition}
    We also define the fine-grained learning rates as defined in \cite{bousquet2022finegraineddistributiondependentlearningcurves}.

\begin{definition}[Marginal-nonuniform fine-grained rates]\label{def:fine-grained}
Let $\H$ be a concept class, let $\lambda \in [0,1]$, and let $R : \N \to (0,\infty)$ be a function with
$\lim_{n\to\infty} R(n) = 0$ that depends only on $\H$ and $\lambda$ (and not on the distribution).
\begin{enumerate}
    \item $\H$ is \emph{$\lambda$-agnostically marginal-nonuniformly learnable at fine-grained rate $R$} if there
    are a universal constant $K > 0$ and a learning algorithm $\A$ such that for every marginal distribution
    $\D_\X$ there is a function $\delta_{\D_\X} : \N \to [0,\infty)$ with
    $\lim_{n\to\infty} \delta_{\D_\X}(n)/R(n) = 0$, such that for every distribution $\D$ with marginal $\D_\X$
    and every $n \in \N$,
    \[
        \E_{S\sim\D^n}\lrbra{\er_\D(\A_S)} - (1+\lambda)\inf_{h\in\H}\er_\D(h) \le K\,R(n) + \delta_{\D_\X}(n).
    \]

    \item $\H$ is \emph{not $\lambda$-agnostically marginal-nonuniformly learnable at fine-grained rate faster
    than $R$} if there is a universal constant $\kappa > 0$ such that for every learning algorithm $\A$ there
    is a marginal distribution $\D_\X$ such that for infinitely many $n \in \N$ there is a distribution
    $\D_n$ with marginal $\D_\X$ satisfying
    \[
        \E_{S\sim\D_n^n}\lrbra{\er_{\D_n}(\A_S)} - (1+\lambda)\inf_{h\in\H}\er_{\D_n}(h) \ge \kappa\,R(n).
    \]

    \item $\H$ is \emph{$\lambda$-agnostically marginal-nonuniformly learnable at exact fine-grained rate $R$}
    if both 1.\ and 2.\ hold.
\end{enumerate}
\end{definition}
    Our main results are based on two combinatorial dimensions; namely, the \emph{Vapnik-Chervonenkis} (VC) dimension as defined in \cite{vc1971} and the {uncentered VC-eluder} dimension essentially defined in \cite{brukhim2026marginal}.
    \begin{definition}
    \label{def:vc}
        \textnormal{\textbf{($\mathbfcal{H}$ VC-shatters $\mathbf{S}$)}}
        We say $\H$ VC-shatters a set $S$ if the projection of $\H$ onto $S$ i.e., $\H|_S$ has size $2^{|S|}$. 
        
        \textnormal{\textbf{(VC dimension)}}
        The \emph{VC dimension} of $\H$, denoted $\VC(\H)$, is the size of the largest set that $\H$ VC-shatters, or $\infty$ if no such largest set exists.
    \end{definition}
    \begin{definition}
    \label{def:vceluder}
        \textnormal{\textbf{(uncentered VC-eluder sequence)}}
        We say that $\H$ has an infinite uncentered $d$-VC-eluder sequence $B_1,B_2,\ldots$ if each $B_k \in \X^d$ is VC-shattered by concepts in $\H$ that are consistent with each other over $\cup_{j < k}B_j$.

        \textnormal{\textbf{(uncentered VC-eluder dimension)}}
        The \emph{uncentered VC-eluder dimension} of $\H$, denoted by $\uVCE(\H)$, is defined as largest $d$ for which $\H$ has an infinite uncentered $d$-VC-eluder sequence. We refer to uncentered $1$-VC-eluder sequences as \emph{infinite u-eluder} sequence. 
        If no infinite sequence exists then $\uVCE(\H)=0$ and $\uVCE(\H)=\infty$ if there exists an infinite $d$-VC-eluder sequence for every $d\in \N$.
    \end{definition}
    The \emph{centered} version of the above dimension can be defined analogously by fixing a concept $h^* \in \Y^\X$ and replacing the consistency condition by the requirement that every concept in the corresponding witnessing family agrees with $h^*$ on all previously selected points. See Appendix \ref{appendix:centeredvsuncentered} and \cite{brukhim2026marginal} for a detailed comparison.

    We now recall a few other standard notions. For every concept class $\H \subseteq \Y^\X$, every finite sequence of instances $\U=(x_1,\ldots,x_n)\in \X^n$ and every concept $h \in \Y^\X$ we define the corresponding \emph{version space} by 
    $\V_{\U,h}= \lrset{f \in \H:f(x_i)=h(x_i) \text{ for all } i \in [n]}$. 
    Similarly, we can do so, if we have a finite sequence of instance-label pairs. In addition, we define the disagreement region of a class $\H$ by $\mathrm{DIS}(\H) = \{x\in \X|\text{ there exists }f,g \in \H \text{ such that } f(x) \neq g(x)\}$. 
    Finally, for every class $\V \subseteq \Y^\X$ and $n \in \N$, let
$\SH(\V,n) = \{(x_1,\ldots,x_n) \in \X^n : \lrset{(h(x_1),\ldots,h(x_n)) : h\in\V} =
$ $\{0,1\}^n\}$
be the set of $n$-tuples that are VC-shattered by $\V$. A tuple with a repeated point is never shattered. 

     \begin{definition}[Distributional VC dimension]
    For $\V \subseteq \H$, a marginal distribution $\D_\X$ and an integer $k \ge 0$, we write
    $\VC_{\D_\X}(\V) \le k$ if $\D_\X^{k+1}\lr{\SH(\V,k+1)} = 0$, i.e., an i.i.d.\ sample of $k+1$ points
    from $\D_\X$ is almost surely not VC-shattered by $\V$.
\end{definition}
If $\VC_{\D_\X}(\V)\le k$, then almost surely the projection of $\V$ onto an i.i.d.\ sample from $\D_\X$ of any
size has VC dimension at most $k$, and the same holds for every subclass of $\V$. In particular
$\VC_{\D_\X}(\V') \le \VC_{\D_\X}(\V)$ whenever $\V' \subseteq \V$.
    Our upper bounds use a first-order learner whose guarantee depends only on the distributional VC dimension; Appendix~\ref{appendix:firstorderlearner} explains why this is needed and shows that such a learner exists.

For $\V \subseteq \H$ write $\tau_{\D,\V} = \inf_{h \in \V}\er_\D(h)$ and $\tau_\D = \tau_{\D,\H}$.

\begin{definition}[First-order learner]\label{def:firstorderlearner}
A \emph{first-order learner} is a (possibly randomized) learning rule $\mathcal B$ that takes a class $\V \subseteq \H$ and a sample $S$, and outputs a classifier $\mathcal B_\V(S)$, such that for a universal constant $B \ge 1$ the following holds. 
For every class $\V$ that is $\H$ or a version space of $\H$, every distribution $\D$ over $\X\times\Y$, every $k \ge 1$ with $\VC_{\D_\X}(\V) \le k$, and every $n \ge 1$,
\begin{equation}\label{eq:first-order}
    \E_{S\sim\D^n}\lrbra{\er_\D(\mathcal B_\V(S))} \le \tau_{\D,\V} + B\lr{\sqrt{\frac{\tau_{\D,\V}\,k}{n}} + \frac{k}{n}}.
\end{equation}
The rule uses only $\V$ and $S$; it does not know $\D_\X$, $k$ or $\tau_{\D,\V}$. 
The expectation includes the internal randomness of $\mathcal B$.
The existence of such a learner has been discussed in Appendix \ref{appendix:firstorderlearner}.
\end{definition}
\section{Standard Agnostic Setting}
In this section we note that in the standard agnostic setting the dependence on the marginal brings no benefit: both the rates and the fine-grained rates are those of uniform agnostic PAC learning. 
The upper bounds are uniform bounds, so they hold in particular in the marginal-nonuniform model. 
We also note that the classical lower bound can be realized with a single marginal i.e. the uniform distribution, so that it also holds in the marginal-nonuniform model. 
We first prove the fine-grained statement and then deduce the rates from it.
\begin{theorem}[Fine-grained rate, standard agnostic]\label{thm:standard-fine}
    Let $\H$ be a concept class with $ d = \VC(\H) < \infty$.
    \begin{enumerate}
        \item \emph{(Upper bound.)} There is a learner $\A$ such that for every distribution $\D$ and every $n \ge 1$,
        \[
            \E_{S\sim\D^n}\lrbra{\er_\D(\A_S)} - \tau_\D \le B\lr{\sqrt{\frac{d}{n}} + \frac{d}{n}},
        \]
        where $B$ is the constant of the first-order learner in Definition \ref{def:firstorderlearner}.
        \item \emph{(Lower bound.)} There is a marginal distribution $\D_\X$ such that for every learner $\A$ and
    every $n \ge d$ there is a distribution $\D$ with marginal $\D_\X$ satisfying
\[
    \E_{S\sim\D^n}\lrbra{\er_\D(\A_S)} - \tau_\D \ge c\sqrt{\frac{d}{n}},
\]
where $c>0$ is a universal constant.

    \end{enumerate}
    In particular, $\H$ is $0$-agnostically marginal-nonuniformly learnable at exact fine-grained rate $\sqrt{\frac{d}{n}}$.
\end{theorem}

\begin{proof}
    \emph{Upper bound.} Run the first-order learner on the whole class, $\A = \mathcal B_\H$. 
    Every marginal $\D_\X$ satisfies $\VC_{\D_\X}(\H) \le \VC(\H) = d$, so \eqref{eq:first-order} applies with $\V = \H$ and $k = d$.
       With $\tau_\D \le 1$, Definition~\ref{def:firstorderlearner} gives the bound.
    The leading term $B\sqrt{d/n}$ depends only on $d$, and the remaining term $Bd/n$ is of lower order, so this is a fine-grained upper bound. 

    \emph{Lower bound.} This is the classical lower bound for agnostic PAC learning \citep[Section 28.2]{shalev2014understanding}. 
    The point is that its proof already works with a single marginal. 
    For any class $\H$ with $\VC(\H) = d$, all the hard distributions used there have the same marginal: the uniform distribution $\D_\X$ on a fixed set $\{x_1,\ldots,x_d\}$ that is VC-shattered by $\H$. 
    Only the conditional distribution of the labels, which is slightly biased towards a labeling of the shattered set, is chosen depending on $n$ and on the learner $\A$. 
    Hence for this one marginal $\D_\X$, for every learner $\A$ and every $n \ge d$, there is a distribution $\D$ with marginal $\D_\X$ whose expected excess error is at least $c\sqrt{d/n}$.
    In the notation of Definition~\ref{def:fine-grained}, the upper bound gives item 1 with $R(n) = \sqrt{d/n}$,
    $K = B$ and $\delta(n) = B\frac{d}{n} = o(R(n))$, and the lower bound gives item 2 with $\kappa = c$, since it
    holds for every $n \ge d$.
\end{proof}

\begin{theorem}[Rates, standard agnostic]\label{thm:standard-rates}
    Let $\H$ be a concept class with $|\H| \ge 3$.
    \begin{enumerate}
        \item $\H$ is $0$-agnostically marginal-nonuniformly learnable at exact rate $\frac{1}{\sqrt{n}}$ if and only if $\VC(\H)<\infty$.
        \item $\H$ requires arbitrarily slow rates to be $0$-agnostically marginal-nonuniformly learnable if and only if $\VC(\H)=\infty$.
    \end{enumerate}
\end{theorem}

\begin{proof}

    Since $|\H| \ge 2$, some point receives two different labels from concepts in $\H$, so $\VC(\H) \ge 1$.

\emph{Finite VC dimension.} Let $d = \VC(\H) < \infty$. By the upper bound of Theorem~\ref{thm:standard-fine},
for $n \ge d$ the excess error is at most $B\lr{\sqrt{d/n} + d/n} \le 2B\sqrt{d/n}$, and for $n < d$ it is at
most $1 \le \sqrt{d/n} \le 2B\sqrt{d/n}$, since $B \ge 1$. So for every $n$ the excess error is at most
$2B\sqrt{d}\cdot(1/\sqrt{n})$, and $\H$ is learnable at rate $1/\sqrt{n}$, with constants that do
not even depend on the marginal. By the lower bound of Theorem~\ref{thm:standard-fine}, a single marginal
forces excess error at least $c\sqrt{d}\cdot(1/\sqrt{n})$ at every $n \ge d$, so $\H$ is not learnable
at a rate faster than $1/\sqrt{n}$. This proves the ``if'' direction of the first item, and it also
shows that $\H$ does not require arbitrarily slow rates.

    \emph{Infinite VC dimension.} Every realizable distribution, i.e., one whose labels are given by some $f \in \H$, is allowed in our model and has $\tau_\D = 0$, so for it the excess error is the error itself. Hence any lower bound for realizable marginal-nonuniform learning is also a lower bound here. 
    If $\VC(\H) = \infty$, the class requires arbitrarily slow rates already in the realizable case \citep[Theorem 3]{hanneke2025marginalnonuniform}, hence also in the $0$-agnostic case. In particular it is not learnable at rate $\frac{1}{\sqrt n}$, which proves the ``only if'' direction of the first item.
\end{proof}

\section{Relaxed Agnostic Setting}

\begin{theorem}[Rates, relaxed agnostic]\label{thm:relaxed-rates}
    Let $\H$ be a concept class with $|\H| \ge 3$ and let $0 < \lambda \le 1$.
    \begin{enumerate}
        \item $\H$ is $\lambda$-agnostically marginal-nonuniformly learnable at exact rate $e^{-n}$ if and only if $\H$
        is finite.
        \item $\H$ is $\lambda$-agnostically marginal-nonuniformly learnable at exact rate $\frac{1}{n}$ if and only if
        $\H$ is infinite and $\VC(\H)<\infty$.
        \item $\H$ requires arbitrarily slow rates to be $\lambda$-agnostically marginal-nonuniformly learnable if and
        only if $\VC(\H)=\infty$.
    \end{enumerate}
\end{theorem}

The three cases are exhaustive and mutually exclusive. For the first two cases, we establish matching upper and lower bounds; the third requires only a lower bound.

\emph{Lower bounds.} As in the standard setting, every realizable distribution has $\tau_\D = 0$, so for it the relaxed excess error $\E\lrbra{\er_\D(\A_S)} - (1+\lambda)\tau_\D$ is the error itself, whatever $\lambda$ is. 
Hence the lower bounds of the realizable trichotomy \citep[Theorem~3]{hanneke2025marginalnonuniform} carry over: a class with $|\H| \ge 3$ is not learnable faster than $e^{-n}$, an infinite class is not learnable faster than $\frac1n$, and a class with $\VC(\H) = \infty$ requires arbitrarily slow rates.

\emph{Upper bounds.} For a finite class we prove an exponential rate in Theorem \ref{thm:relaxed-finite} below. For an
infinite class with finite VC dimension, Lemma \ref{lem:uvce-basic2} gives $1 \le \uVCE(\H) < \infty$, and the upper bound
of Theorem \ref{thm:relaxed} gives excess error at most $K\frac{\uVCE(\H)}{\lambda n} + Ce^{-cn}$, which is a rate
$\frac1n$.
\subsection{Finite classes}

For a finite class, a plain ERM already achieves an exponential rate. 
The idea is that ERM compares two concepts only on the points where they disagree. 
On the disagreement region of the best concept and a concept with error more than $(1+\lambda)\tau_\D$, the best concept is correct noticeably more often than not, and for a fixed marginal this region has probability bounded away from $0$. 
So ERM prefers the worse concept only with exponentially small probability. 
This uses $\lambda > 0$ in an essential way: under the standard criterion even a class of two concepts has rate $\frac{1}{\sqrt n}$.

\begin{theorem}[Exponential rate for finite classes]\label{thm:relaxed-finite}
    Let $\H$ be a finite concept class and $0 < \lambda \le 1$. Let $\A$ be any ERM over $\H$. For every marginal $\D_\X$, let
    \[
        \delta_{\D_\X} = \min\lrset{\D_\X\lr{\DIS(\{h,g\})} : h, g \in \H,\ \D_\X\lr{\DIS(\{h,g\})} > 0},
    \]
    with $\delta_{\D_\X} = 1$ if this set is empty. Then for every distribution $\D$ with marginal $\D_\X$ and every $n \ge 1$,
    \[
        \E_{S\sim\D^n}\lrbra{\er_\D(\A_S)} - (1+\lambda)\tau_\D \le |\H|\exp\lr{-\frac{\lambda^2\delta_{\D_\X}}{19}\,n}.
    \]
\end{theorem}

The constants $|\H|$ and $\frac{\lambda^2\delta_{\D_\X}}{19}$ depend on $\H$, $\lambda$ and $\D_\X$ but not on the conditional distribution of the labels, so this is an exponential marginal-nonuniform rate.

\begin{proof}
    Fix $\D$ with marginal $\D_\X$, and let $h^* \in \H$ have $\er_\D(h^*) = \tau_\D$; it exists because $\H$ is finite. 
    Call a concept $h \in \H$ \emph{bad} if $\er_\D(h) > (1+\lambda)\tau_\D$. If ERM outputs a concept that is not bad, its error is at most $(1+\lambda)\tau_\D$; otherwise we bound its error by $1$. 
    Hence
    \begin{equation}\label{eq:finite-split}
        \E\lrbra{\er_\D(\A_S)} \le (1+\lambda)\tau_\D + \P\lrbra{\A_S \text{ is bad}},
    \end{equation}
    and it remains to show that ERM rarely outputs a bad concept. Fix a bad concept $h$.

    \textbf{Only the disagreement region matters.} 
    Let $R = \DIS(\{h,h^*\})$ be the set of points where $h$ and $h^*$ disagree, and let $s = \D_\X(R)$. 
    Outside $R$ the two concepts make exactly the same mistakes. 
    Inside $R$ the labels are binary, so exactly one of them is wrong: $h$ makes a mistake exactly when $h^*$ is correct. Let $q = \P_{(X,Y)\sim\D}\lrbra{h^*(X) \ne Y \mid X \in R}$ be the probability that $h^*$ errs on a point of $R$
    (this is well defined, since $s > 0$ as we show next). 
    Then
    \[
        \er_\D(h) - \er_\D(h^*) = s(1-q) - sq = s(1-2q).
    \]

    \textbf{On the disagreement region, $h^*$ is right noticeably more often than wrong.} 
    Since $h$ is bad, $\er_\D(h) - \er_\D(h^*) > \lambda\tau_\D \ge 0$, so $s > 0$. 
    Also $\tau_\D = \er_\D(h^*) \ge sq$, because the mistakes of $h^*$ on $R$ are part of all its mistakes. 
    Hence $s(1-2q) > \lambda sq$, i.e., $q < \frac{1}{2+\lambda}$, and with $\lambda \le 1$,
    \begin{equation}\label{eq:finite-margin}
        \frac12 - q > \frac12 - \frac{1}{2+\lambda} = \frac{\lambda}{2(2+\lambda)} \ge \frac{\lambda}{6}.
    \end{equation}
    Moreover, $s > 0$ means that the set in the definition of $\delta_{\D_\X}$ is nonempty and contains $s$, so $s \ge \delta_{\D_\X}$.

    \textbf{ERM rarely prefers $h$ to $h^*$.} 
    Let $N$ be the number of sample points that fall in $R$, and let $M$ be the number of those at which $h^*$ makes a mistake. 
    On the sample, outside $R$ the two concepts have the same mistakes, and inside $R$, $h^*$ makes $M$ mistakes and $h$ makes $N - M$. ERM can output $h$ only if $h$ makes at most as many mistakes on $S$ as $h^*$, that is, only if $M \ge N/2$. 
    We bound this probability in two steps.
    \begin{itemize}
        \item \emph{Given $N$.} Conditioned on which sample points fall in $R$, these points are i.i.d. draws from $\D$ conditioned on $X \in R$, so $M$ is a sum of $N$ independent indicators with mean $q$. 
        By Hoeffding's inequality and \eqref{eq:finite-margin},
        \[
            \P\lrbra{M \ge \tfrac N2 \,\middle|\, N} \le \exp\lr{-2N\lr{\tfrac12 - q}^2} \le e^{-tN},\text{where } t = \frac{\lambda^2}{18}.
        \]
        This also holds when $N = 0$, where the right side equals $1$.
        \item \emph{Averaging over $N$.} Each sample point falls in $R$ independently with probability $s$, so $N$ is binomial with parameters $n$ and $s$. Hence, using $1-u \le e^{-u}$,
        \[
            \E\lrbra{e^{-tN}} = \lr{1 - s\lr{1 - e^{-t}}}^n \le \exp\lr{-sn\lr{1-e^{-t}}}.
        \]
        Since $t \le \frac1{18}$, we have $1 - e^{-t} \ge \frac{t}{1+t} \ge \frac{18}{19}t = \frac{\lambda^2}{19}$.
    \end{itemize}
    Combining the two steps and $s \ge \delta_{\D_\X}$,
    \[
        \P\lrbra{\A_S = h} \le \P\lrbra{M \ge \tfrac N2} \le \exp\lr{-\frac{\lambda^2 s}{19}\,n} \le \exp\lr{-\frac{\lambda^2 \delta_{\D_\X}}{19}\,n}.
    \]

    \textbf{Union bound.} There are at most $|\H|$ bad concepts, so $\P\lrbra{\A_S \text{ is bad}} \le |\H|\exp\lr{-\lambda^2\delta_{\D_\X}n/19}$. 
    Together with \eqref{eq:finite-split} this proves the theorem. (If no concept is bad, the bound holds trivially.)
\end{proof}
\subsection{Fine-grained rates}

We now turn to the fine-grained rates for infinite classes of finite VC dimension. 
We state the result here and prove the upper and lower bounds in the next two subsections.
\begin{theorem}[Fine-grained rate, relaxed agnostic]\label{thm:relaxed}\label{thm:relaxed-fine}
Let $0 < \lambda \le 1$ and $1 \le \uVCE(\H) < \infty$. Then $\H$ is $\lambda$-agnostically
marginal-nonuniformly learnable at exact fine-grained rate $\frac{\uVCE(\H)}{\lambda n}$.
\end{theorem}
Theorem \ref{thm:relaxed} follows from Theorems \ref{thm:relaxed-ub} and \ref{thm:relaxed-lb}, with
$R(n) = \frac{\uVCE(\H)}{\lambda n}$, which depends only on $\H$ and $\lambda$. The upper bound gives item 1
of Definition~\ref{def:fine-grained} with the constant $K$ of Theorem~\ref{thm:relaxed-ub}, which depends only
on the universal constant $B$ of the first-order learner, and with $\delta_{\D_\X}(n) = Ce^{-cn}$, which is
$o(R(n))$. The lower bound gives item 2 with $\kappa = \frac{1}{2048}$, along the infinite sequence
$n_1 < n_2 < \cdots$ of Theorem~\ref{thm:relaxed-lb}.
\subsection{Fine-grained upper bound}
\label{sec:upper}
\begin{theorem}\label{thm:relaxed-ub}
    Let $0 < \lambda \le 1$ and suppose $1 \le d = \uVCE(\H) < \infty$. There are a learner $\A_\lambda$ and a
    constant $K$ depending only on $B$ from the first-order learner as in Definition \ref{def:firstorderlearner} such that for every marginal $\D_\X$ there are $C, c > 0$, depending on
    $\D_\X$, $\lambda$ and $\H$ only, with
    \[
        \E_{S \sim \D^n}\lrbra{\er_\D(\A_{\lambda,S})} - (1+\lambda)\tau_\D \le K\,\frac{d}{\lambda n} + C e^{-cn}
    \]
    for every $\D$ with marginal $\D_\X$ and every $n \ge 1$.
\end{theorem}
\paragraph{Idea.} In the realizable case \cite{brukhim2026marginal}, the learner uses part of the sample to form a version space. 
After enough samples this version space is ``simple'': a fresh sample from $\D_\X$ almost never contains $d+1$ points that it shatters. 
A good learner is then run inside this simple class.
In the agnostic case the sample is not consistent with any single concept, so we cannot form this version space. 
However, a good concept $h$ makes few mistakes, so a small random subsample is likely to contain none of them. 
The version space of such a \emph{clean} subsample is again simple, and it still contains $h$.
We do not know which subsamples are clean, so we try all of them: every subsample gives one candidate predictor, obtained by running the first-order learner inside its version space. 
Finally, we choose among the candidates on fresh data using exponential weights. 
The factor $(1+\lambda)$ is what pays for this last selection step.

\subsubsection{The learner}

Fix $\eta = \frac{\lambda}{16}$ and $p = \frac{\lambda^2}{1024}$. 
For $n < 6$ the learner outputs a fixed $h \in \H$. For $n \ge 6$ let $m = \lfloor n/3 \rfloor$ and $r = n - 2m$, so that $\frac n4 \le m \le r$, and split $S$ into three independent parts, one for each stage of the idea above:
\[
    \underbrace{S^{(0)} = ((X_i,Y_i))_{i=1}^m}_{\text{to localize}}, \qquad
    \underbrace{S^{(1)} \text{ of size } m}_{\text{to train the candidates}}, \qquad
    \underbrace{S^{(2)} \text{ of size } r}_{\text{to choose among them}}.
\]
\begin{enumerate}
    \item \emph{Candidates (experts).} Let
    \[
        \mathcal E = \{\star\} \cup \lrset{I \subseteq [m] : (X_i,Y_i)_{i\in I} \text{ is realizable by } \H}.
    \]
    For $I \in \mathcal E\setminus\{\star\}$, form the version space $\V_I = \V_{(X_i,Y_i)_{i\in I}}$ and train
    the first-order learner inside it, $g_I = \mathcal B_{\V_I}(S^{(1)})$. For the special expert $\star$, let
    $g_\star$ be an ERM over $\H$ on $S^{(1)}$. The experts are the elements of $\mathcal E$: two different
    experts may output the same classifier, and we still treat them as different experts. This use of realizable subsamples follows the agnostic-to-realizable reductions of \citet{david2016supervised}, \citet{hopkins2022realizable}, and \citet{hanneke2025representation}.

    \item \emph{Prior.} Let $w(I) = p^{|I|}(1-p)^{m-|I|}$. This is the probability that a random subset of
    $[m]$, which contains each index independently with probability $p$, is exactly $I$. Let
    $Z = \sum_{I \in \mathcal E\setminus\{\star\}} w(I)$. The empty subsample is realizable, so
    $Z \ge w(\emptyset) = (1-p)^m > 0$, and $Z \le 1$. Set
    \[
        \pi(\star) = \frac12, \qquad \pi(I) = \frac{w(I)}{2Z} \ge \frac{w(I)}{2} \quad \text{for } I \in \mathcal E\setminus\{\star\}.
    \]
    So $\pi$ is a probability distribution on $\mathcal E$ that favors small subsamples.

    \item \emph{Aggregation.} Let $(X'_t,Y'_t)$, $t\in[r]$, be the examples of $S^{(2)}$, and for $e \in \mathcal E$
    let $\ell_t(e) \in [0,1]$ be the probability that $g_e$ errs on $(X'_t,Y'_t)$ (over the internal
    randomness of $g_e$, if any). Run exponential weights with rate $\eta$ starting from $\pi$:
    \[
        w_1 = \pi, \qquad w_{t+1}(e) = \frac{w_t(e)\,e^{-\eta\ell_t(e)}}{\sum_{e'\in\mathcal E} w_t(e')\,e^{-\eta\ell_t(e')}} .
    \]
    Output the randomized classifier $\hat h$ that draws $t \in [r]$ uniformly, then an expert $e \in \mathcal E$
    from $w_t$, and predicts $g_e(x)$. Its error is $\er_\D(\hat h) = \frac1r\sum_{t=1}^r\sum_{e\in\mathcal E} w_t(e)\,\er_\D(g_e)$.
\end{enumerate}

\subsubsection{Version spaces become simple}

The first tool is taken from \cite{brukhim2026marginal}: after enough samples from $\D_\X$, all the version spaces are simple at the same time.

\begin{lemma}[Exhaustion]\label{lem:exhaustion}
    If $\H$ has no infinite uncentered $(d+1)$-VC-eluder sequence, then for every marginal $\D_\X$,
    \[
        \lim_{m\to\infty}\P_{\U \sim \D_\X^m}\lrbra{\text{there is } h \in \H \text{ with } \VC_{\D_\X}(\V_{\U,h}) > d} = 0.
    \]
\end{lemma}
This is Corollary 20 of \cite{brukhim2026marginal}, stated at width $d+1$. 
We include the proof, which follows their Lemma 18.
\begin{proof}
Suppose, towards a contradiction, that the limit is not $0$ for some $\D_\X$. Let $X_1, X_2, \ldots$ be
i.i.d.\ from $\D_\X$ and write $\U_m = (X_1,\ldots,X_m)$, with $\U_0$ empty.

\textbf{Version spaces depend only on labels.} For $\sigma \in \{0,1\}^m$ let
$\V_{\U_m,\sigma} = \{f\in\H : f(X_i) = \sigma_i$ for all  $i\in[m]\}$. For every $h\in\H$ we have
$\V_{\U_m,h} = \V_{\U_m,\sigma}$ with $\sigma = (h(X_1),\ldots,h(X_m))$. So, although $\H$ may be
uncountable, there are at most $2^m$ different version spaces on $\U_m$.

\textbf{A bad event that never goes away.} Let $E_m$ be the event that some $\sigma\in\{0,1\}^m$ satisfies
$\V_{\U_m,\sigma} \ne \emptyset$ and $\D_\X^{d+1}\lr{\SH(\V_{\U_m,\sigma},d+1)} > 0$. This is exactly the event
that some $h\in\H$ has $\VC_{\D_\X}(\V_{\U_m,h}) > d$, and it is a finite union of measurable events. If
$\sigma'\in\{0,1\}^{m+1}$ witnesses $E_{m+1}$, then its first $m$ coordinates $\sigma$ witness $E_m$, because
$\V_{\U_{m+1},\sigma'} \subseteq \V_{\U_m,\sigma}$ and a smaller class shatters fewer tuples. Hence
$E_{m+1}\subseteq E_m$. By assumption $\lim_m \P[E_m] > 0$, so $E_\infty = \bigcap_m E_m$ has positive
probability.

\textbf{A set of shattered tuples of positive mass.} On $E_m$, let $\sigma_m$ be the lexicographically
first $\sigma\in\{0,1\}^m$ that witnesses $E_m$, and let $A_m = \SH(\V_{\U_m,\sigma_m},d+1)$, so that
$\D_\X^{d+1}(A_m) > 0$. Off $E_m$, let $A_m = \emptyset$. Since $\sigma_m$ is chosen among finitely many
candidates by a fixed rule, $A_m$ is determined by $\U_m$ in a measurable way; we never choose a concept
from $\H$.

\textbf{Future samples hit this set.} Cut the future samples $X_{m+1}, X_{m+2},\ldots$ into consecutive
disjoint blocks of $d+1$ points. These blocks are i.i.d.\ from $\D_\X^{d+1}$ and independent of $\U_m$.
Conditioned on $\U_m$, each block lies in $A_m$ independently with probability $\D_\X^{d+1}(A_m)$, which
is positive on $E_m$. By the second Borel--Cantelli lemma, almost surely on $E_m$ infinitely many blocks
lie in $A_m$. There are countably many $m$, so almost surely this holds for all $m$ at once.

\textbf{Build the forbidden sequence.} Fix a sample path in $E_\infty$ on which the conclusion of the
previous step holds for all $m$; the rest is deterministic. Set $t_0 = 0$. Given $t_{r-1}$, pick a block
after position $t_{r-1}$ that lies in $A_{t_{r-1}}$, call its set of points $B_r$, and let $t_r$ be the
position of its last point. By definition of $A_{t_{r-1}}$, the set $B_r$ is VC-shattered by
$\V_{\U_{t_{r-1}},\sigma_{t_{r-1}}}$. All members of this version space have the labels $\sigma_{t_{r-1}}$
on $\U_{t_{r-1}}$, so they agree with each other there, and $\U_{t_{r-1}}$ contains
$B_1,\ldots,B_{r-1}$. Hence $B_1, B_2,\ldots$ is an infinite uncentered $(d+1)$-VC-eluder sequence, a
contradiction.
\end{proof}

Since $\uVCE(\H) = d$, the class has no infinite uncentered $(d+1)$-VC-eluder sequence and Lemma \ref{lem:exhaustion} applies. 
\emph{For the rest of this section fix a marginal $\D_\X$.} 
By the lemma there is $q$ which depends on $\D_\X$ such that
with
\begin{equation}\label{eq:good-block}
    \P_{\U \sim \D_\X^q}\lrbra{\VC_{\D_\X}(\V_{\U,h}) \le d \text{ for every } h \in \H} \ge \frac34 .
\end{equation}
We call a $q$-tuple $\U$ \emph{good} if the event in \eqref{eq:good-block} holds, and we set
\[
    \rho = \frac{1}{4q}, \qquad \beta = \frac{3p}{4}\lr{1 - 2^{-1/q}} > 0 .
\]
These quantities are used only in the analysis; the learner does not know $q$, $\rho$ or $\beta$.

\subsubsection{Clean subsamples localize}

Lemma \ref{lem:exhaustion} is about samples from $\D_\X$. 
The next lemma transfers it to what the learner can actually use: a random subsample on which a fixed good concept $h$ makes no mistake.

\begin{lemma}[Localization from a clean subsample]\label{lem:localization}
    Let $h \in \H$ with $a = \er_\D(h) \le \rho$. Draw $S^{(0)} \sim \D^m$ and independent $b_1,\ldots,b_m \sim
    \mathrm{Ber}(p)$. Keep the indices that are selected and on which $h$ is correct,
    \[
        I = \lrset{i \in [m] : b_i = 1 \text{ and } Y_i = h(X_i)}, \qquad \U_I = (X_i)_{i \in I}.
    \]
    Then
    \[
        \P\lrbra{\VC_{\D_\X}(\V_{\U_I,h}) > d} \le 2e^{-\beta m}.
    \]
\end{lemma}
\begin{proof}

    \textbf{The kept points are i.i.d..} The triples $(X_i,Y_i,b_i)$ are i.i.d., and index $i$ is kept
    with probability $p(1-a)$. 
    A kept point has the law $\nu_h$ of $X$ conditioned on $Y = h(X)$. 
    Hence the number of kept points is $J = |I| \sim \mathrm{Bin}(m, p(1-a))$, and given $J$ the kept points are i.i.d. from $\nu_h$.

    \textbf{$\boldsymbol{\nu_h}$ is close to $\mathbfcal{D_X}$.} Conditioning on whether $h$ is correct gives $\D_\X = (1-a)\nu_h + a\nu'$, where $\nu'$ is the law of $X$ given $Y \ne h(X)$. 
    So $\mathrm{TV}(\nu_h,\D_\X) \le a$, and for $q$-tuples $\mathrm{TV}(\nu_h^q,\D_\X^q) \le qa \le q\rho = \frac14$.
    By \eqref{eq:good-block}, a $q$-tuple drawn from $\nu_h^q$ is therefore good with probability at least $\frac34 - \frac14 = \frac12$.

    \textbf{One good block is enough.} Split the kept points into $\lfloor J/q \rfloor$ disjoint blocks of
    $q$ points. If some block $\U'$ is good, then $\VC_{\D_\X}(\V_{\U',h}) \le d$, and since
    $\V_{\U_I,h} \subseteq \V_{\U',h}$ also $\VC_{\D_\X}(\V_{\U_I,h}) \le d$. The blocks are independent and each
    fails to be good with probability at most $\frac12$, so
    \[
        \P\lrbra{\VC_{\D_\X}(\V_{\U_I,h}) > d \mid J} \le \lr{\frac12}^{\lfloor J/q\rfloor} \le 2\cdot 2^{-J/q}.
    \]

    \textbf{Average over $\mathbf J$.} For $J \sim \mathrm{Bin}(m,p')$ and $x \in [0,1]$ we have
    $\E\lrbra{x^J} = (1 - p'(1-x))^m \le e^{-mp'(1-x)}$. With $p' = p(1-a) \ge \frac34p$ (as $a \le \rho \le \frac14$)
    and $x = 2^{-1/q}$ this gives $\E\lrbra{2^{-J/q}} \le e^{-\beta m}$, and the claim follows from the previous step.
\end{proof}

\subsubsection{Choosing among the candidates}

The next lemma is the standard guarantee of exponential weights \citep{freund1997decision}, stated against
a distribution $Q$ over experts instead of a single expert and combined with the usual online-to-batch
conversion \citep{cesabianchi2004generalization}.
The price for competing with $Q$ is $\KL(Q\|\pi)$, which is small whenever the prior puts enough mass where $Q$ lives.

\begin{lemma}[Exponential weights against a distribution]\label{lem:mw}
Let $a_\eta = 1 - e^{-\eta}$ and $b_\eta = \eta/a_\eta$. For every distribution $Q$ on $\mathcal E$ that is
determined by $S^{(0)}$ and $S^{(1)}$,
\[
    \E\lrbra{\er_\D(\hat h) \,\middle|\, S^{(0)}, S^{(1)}} \le b_\eta \sum_{e\in\mathcal E} Q(e)\,\er_\D(g_e) + \frac{\KL(Q\|\pi)}{r\,a_\eta},
\]
where $\log$ is the natural logarithm, and for a randomized expert, $\er_\D(g_e)$ includes its internal
randomness. If $Q = \pi(\cdot \mid G)$ for a set $G \subseteq \mathcal E$ with $\pi(G) > 0$, then
$\KL(Q\|\pi) = \log\frac{1}{\pi(G)}$.
\end{lemma}

\begin{proof}
    Condition on $S^{(0)}$ and $S^{(1)}$, so that the experts, $\pi$ and $Q$ are fixed.
    Let $L_e = \sum_{t=1}^r\ell_t(e)$ be the total loss of expert $e$, and consider the potential
    $W_t = \sum_{e\in\mathcal E} \pi(e)\exp\lr{-\eta\sum_{s<t}\ell_s(e)}$, so that $W_1 = 1$ and
    $w_t(e) = \pi(e)\exp\lr{-\eta\sum_{s<t}\ell_s(e)}/W_t$.

    \textbf{The potential drops when the mixture makes a mistake.} Using $e^{-\eta x} \le 1 - a_\eta x$ for
    $x \in [0,1]$ and $1 - u \le e^{-u}$,
    \[
        \frac{W_{t+1}}{W_t} = \sum_{e\in\mathcal E} w_t(e)\,e^{-\eta\ell_t(e)} \le 1 - a_\eta\langle w_t,\ell_t\rangle \le e^{-a_\eta\langle w_t,\ell_t\rangle},
    \]
    so
    \[
        \log W_{r+1} \le -a_\eta\sum_{t=1}^r\langle w_t,\ell_t\rangle .
    \]

    \textbf{The potential stays large if $Q$ is good.} By Jensen's inequality,
    \[
        \log W_{r+1} = \log\sum_{e\in\mathcal E} \pi(e)e^{-\eta L_e} \ge \sum_{e\in\mathcal E} Q(e)\log\frac{\pi(e)e^{-\eta L_e}}{Q(e)}
        = -\KL(Q\|\pi) - \eta\sum_{e\in\mathcal E} Q(e)L_e ,
    \]
    where the sums over $e$ with $Q(e) = 0$ are dropped.

    \textbf{Combine and take expectations.} The previous two steps give
    $a_\eta\sum_t\langle w_t,\ell_t\rangle \le \eta\sum_{e\in\mathcal E} Q(e)L_e + \KL(Q\|\pi)$.
    The weights $w_t$ depend only on $S^{(0)}$, $S^{(1)}$ and the first $t-1$ examples of $S^{(2)}$, and the
    $t$-th example of $S^{(2)}$ is a fresh draw from $\D$, independent of these. So, conditioned on
    $S^{(0)}$ and $S^{(1)}$, the expectation of $\langle w_t,\ell_t\rangle$ is the expectation of
    $\sum_{e\in\mathcal E} w_t(e)\,\er_\D(g_e)$, and the expectation of $\ell_t(e)$ is $\er_\D(g_e)$.
    Summing over $t$, the left side becomes $a_\eta\,r\,\E\lrbra{\er_\D(\hat h) \mid S^{(0)}, S^{(1)}}$
    and the right side becomes $\eta\,r\sum_{e\in\mathcal E} Q(e)\er_\D(g_e) + \KL(Q\|\pi)$.
    Divide by $r\,a_\eta$.

    For $Q = \pi(\cdot\mid G)$ with $\pi(G) > 0$, the ratio $Q(e)/\pi(e)$ equals $\frac{1}{\pi(G)}$ on $G$
    and $Q(e) = 0$ off $G$, which gives the last claim.
\end{proof}

Finally, when $\tau_\D$ is not small we will not need localization at all; a plain ERM is good enough.
\begin{lemma}[ERM at fixed accuracy]\label{lem:erm-fixed}
Suppose $v = \VC(\H) < \infty$. For every $\zeta \in (0,1]$ there are $A, b > 0$, depending only on $v$ and
$\zeta$, such that for every distribution $\D$ and every $m \ge 1$, an ERM $g_\star$ over $\H$ on
$S \sim \D^m$ satisfies
\[
    \E_{S\sim\D^m}\lrbra{\er_\D(g_\star)} \le \tau_\D + \zeta + Ae^{-bm}.
\]
\end{lemma}

\begin{proof}
Write $\er_S(h)$ for the fraction of examples in $S$ that $h$ misclassifies.

\emph{Uniform convergence.} By the VC inequality \citep[Theorem~12.5]{devroye1996} and Sauer's lemma,
for $m \ge v$,
\[
    \P\lrbra{\sup_{h\in\H}\,\lrabs{\er_\D(h) - \er_S(h)} > \frac{\zeta}{2}}
    \le 8\lr{\frac{em}{v}}^{v} e^{-m\zeta^2/128}.
\]

\emph{ERM is accurate outside this event.} If $\lrabs{\er_\D(h) - \er_S(h)} \le \frac{\zeta}{2}$ for every
$h\in\H$, then, since $g_\star$ minimizes the empirical error, for every $h \in \H$
\[
    \er_\D(g_\star) \le \er_S(g_\star) + \frac{\zeta}{2} \le \er_S(h) + \frac{\zeta}{2} \le \er_\D(h) + \zeta .
\]
Taking the infimum over $h\in\H$ gives $\er_\D(g_\star) \le \tau_\D + \zeta$. On the complementary event
we bound $\er_\D(g_\star) \le 1$. Hence, for $m \ge v$,
\[
    \E\lrbra{\er_\D(g_\star)} \le \tau_\D + \zeta + 8\lr{\frac{em}{v}}^{v} e^{-m\zeta^2/128}.
\]

\emph{Absorbing the polynomial factor.} Let $b = \zeta^2/256$. The function
$m \mapsto 8(em/v)^v e^{-m\zeta^2/256}$ is bounded on $[v,\infty)$; let $A_1$ be its supremum. Then for
$m \ge v$ the last term is at most $A_1 e^{-bm}$. For $m < v$ we use the trivial bound
$\E\lrbra{\er_\D(g_\star)} \le 1 \le e^{bv}e^{-bm}$. The lemma follows with $A = \max\{A_1, e^{bv}\}$,
which depends only on $v$ and $\zeta$.
\end{proof}

\subsubsection{Proof of the upper bound}

\begin{proof}[Proof of Theorem \ref{thm:relaxed-ub}]
    Fix a distribution $\D$ with marginal $\D_\X$. 
    For $n < 6$ the learner outputs a fixed concept and the bound holds trivially, since
$K\frac{d}{\lambda n} \ge \frac{K}{6} \ge 1$. We assume $n \ge 6$ from now on. We will use the elementary bounds
    \begin{equation}\label{eq:elementary}
        b_\eta \le 1 + \eta, \qquad a_\eta \ge \frac\eta2, \qquad \log\frac{1}{1-p} \le 2p, \qquad \frac n4 \le m \le r .
    \end{equation}
    The proof has two cases. If the best concept in $\H$ has small error, we compete with the candidates trained in clean version spaces (Case 1). Otherwise the ERM candidate $g_\star$ already suffices (Case 2).

    \textbf{Case 1: $\tau_\D \le \rho/2$.}

    \textbf{A comparator.} 
    Fix any $h \in \H$ with $a = \er_\D(h) \le \rho$; such $h$ exist because
    $\tau_\D < \rho$. Let
    \[
        E_h = \lrset{i \le m : Y_i \ne h(X_i)}, \qquad K_h = |E_h|,
    \]
    be the positions of the mistakes of $h$ on $S^{(0)}$ and their number, so $\E [K_h] = ma$. A subsample $I$ is
    \emph{clean} if $I \cap E_h = \emptyset$. 
    For a clean $I$ the labels on $I$ are those of $h$, so $I \in \mathcal E$, and
    \begin{equation}\label{eq:clean-vs}
        \V_I = \V_{\U_I,h} \ni h .
    \end{equation}

    \textbf{The good experts.} Define
    \[
       G_h = \lrset{I \in \mathcal E\setminus\{\star\} : I \text{ is clean and } \VC_{\D_\X}(\V_{\U_I,h}) \le d}.
   \]
   These are the experts trained in a version space that is simple and contains $h$. The set
   $G_h \subseteq \mathcal E$ depends on $S^{(0)}$ and $h$ only.
    We show that every good expert is accurate and that the good experts together have enough prior mass; exponential weights then finds them in the following steps.

    \textbf{Each good expert is accurate.} Let $I \in G_h$. By \eqref{eq:clean-vs}, $\V_I$ is a version space
of $\H$ with $\tau_{\D,\V_I} \le a$ and $\VC_{\D_\X}(\V_I) \le d$. Since $S^{(1)}$ is independent of
$S^{(0)}$, we can apply Definition~\ref{def:firstorderlearner} to the fixed class $\V_I$ with $k = d$. Its
right-hand side is increasing in $\tau_{\D,\V_I}$, so writing $z = d/m$,
\[
    \E\lrbra{\er_\D(g_I) \,\middle|\, S^{(0)}} \le a + B\sqrt{az} + Bz,
\]
where the expectation is over $S^{(1)}$ and the internal randomness of $\mathcal B$.
    To separate $a$ from $z$ we use $B\sqrt{az} \le \eps a + \frac{B^2z}{4\eps}$ with $\eps = \frac{\lambda}{16}$
    \begin{equation}\label{eq:good-expert}
        \E\lrbra{\er_\D(g_I) \mid S^{(0)}} \le (1+\eps)a + \frac{(B + B^2)z}{\eps}.
    \end{equation}
    This is the only place where the first-order learner is used, and it is why we need a first-order bound: a
    term $\sqrt{z}$ in place of $\sqrt{az}$ could not be absorbed into $\eps a$.

    \textbf{The good experts have enough prior mass.} Recall that $w$ is the law of a random subset $R \subseteq [m]$ that contains each index independently with probability $p$. 
    This subset is clean with probability $(1-p)^{K_h}$, so $\sum_{I \text{ clean}} w(I) = (1-p)^{K_h}$. 
    Let
    \[
        b_h = \P\lrbra{\VC_{\D_\X}(\V_{\U_R,h}) > d \mid S^{(0)},\ R \text{ is clean}}
    \]
    be the fraction of the clean mass that is not good. 
    Then, using $\pi(I) \ge \frac12w(I)$,
    \begin{equation}
    \label{eq:prior-mass}
    \pi(G_h) = \sum_{I\in G_h}\pi(I) \ge \frac12\sum_{I\in G_h} w(I) = \frac12(1-p)^{K_h}(1-b_h).
    \end{equation}
    Conditioned on $S^{(0)}$ and on $R$ being clean, $R$ has the same law as the set $I$ of
Lemma~\ref{lem:localization} given $S^{(0)}$: each index outside $E_h$ is included independently with probability
$p$. 

Hence $b_h = \P\lrbra{\VC_{\D_\X}(\V_{\U_I,h}) > d \mid S^{(0)}}$, and taking expectations over
$S^{(0)}$, Lemma~\ref{lem:localization} gives $\E[b_h] \le 2e^{-\beta m}$.
   By Markov's inequality, the event $T_h = \lrset{b_h \le \tfrac12}$ satisfies $\P\lrbra{T_h^c} \le 4e^{-\beta m}$.
    On $T_h$, \eqref{eq:prior-mass} gives $\pi(G_h) \ge \frac14(1-p)^{K_h}$. 
    In particular $\pi(G_h) > 0$, so $Q = \pi(\cdot\mid G_h)$ is well defined.
    The factor $(1-p)^{K_h}$ is the reason for the choice of prior: every mistake of $h$ costs only a factor $(1-p)$, so the cost below is proportional to the error of $h$.

    \textbf{Aggregate.} The event $T_h$ and the distribution $Q = \pi(\cdot\mid G_h)$ depend on $S^{(0)}$ only. 
    On $T_h$ we apply Lemma \ref{lem:mw} with this $Q$; by the previous steps and \eqref{eq:elementary},
    \[
        \KL(Q\|\pi) = \log\frac{1}{\pi(G_h)} \le \log 4 + K_h\log\frac{1}{1-p} \le \log 4 + 2pK_h .
    \]
    On $T_h^c$ we use the trivial bound $\er_\D(\hat h) \le 1$. 
    Taking expectations, using \eqref{eq:good-expert} for every expert in the support of $Q$, and $\E [K_h] = ma$,
    \begin{equation}\label{eq:case1-raw}
        \E\lrbra{\er_\D(\hat h)} \le b_\eta\lr{(1+\eps)a + \frac{(B+B^2)z}{\eps}} + \frac{\log 4 + 2pma}{r\,a_\eta} + 4e^{-\beta m}.
    \end{equation}

    \textbf{Simplify.} We collect the terms of \eqref{eq:case1-raw} using \eqref{eq:elementary},
    $\eta = \eps = \frac{\lambda}{16}$ and $p = \frac{\lambda^2}{1024}$.
    \begin{itemize}
        \item \emph{Terms with $a$.} Their coefficient is at most
        \[
            (1+\eta)(1+\eps) + \frac{2pm}{r}\cdot\frac{2}{\eta} \le \lr{1+\frac{\lambda}{16}}^2 + \frac{4p}{\eta}
            = \lr{1+\frac{\lambda}{16}}^2 + \frac{\lambda}{16} \le 1 + \frac{\lambda}{4}.
        \]
        \item \emph{Terms of order $\frac1n$.} Since $m, r \ge \frac n4$,
        \[
            b_\eta\frac{(B+B^2)z}{\eps} + \frac{\log 4}{r\,a_\eta}
            \le \frac{17}{16}\cdot\frac{16(B+B^2)}{\lambda}\cdot\frac{4d}{n} + \frac{128\log 4}{\lambda n}
            \le K\,\frac{d}{\lambda n},
        \]
        where $K = 68(B+B^2) + 178$.
        \item \emph{Exponential term.} $4e^{-\beta m} \le 4e^{-\beta n/4}$.
    \end{itemize}
    Therefore,
    \begin{equation}\label{eq:case1-final}
        \E\lrbra{\er_\D(\hat h)} \le \lr{1+\frac{\lambda}{4}}\er_\D(h) + K\frac{d}{\lambda n} + 4e^{-\beta n/4}.
    \end{equation}

    \textbf{Remove the comparator.} The learner does not depend on $h$, and \eqref{eq:case1-final} holds
    for every $h \in \H$ with $\er_\D(h) \le \rho$. Taking such $h$ with $\er_\D(h)$ arbitrarily close to
    $\tau_\D$ gives
    \[
        \E\lrbra{\er_\D(\hat h)} \le (1+\lambda)\tau_\D + K\frac{d}{\lambda n} + 4e^{-\beta n/4}.
    \]

    \textbf{Case 2: $\tau_\D > \rho/2$.}
Now $\tau_\D$ is bounded below by a constant, so an additive error that is a small constant can be charged
to $\lambda\tau_\D$. Since $\uVCE(\H) < \infty$ we have $v = \VC(\H) < \infty$ by Lemma \ref{lem:uvce-basic2}.
Apply Lemma \ref{lem:erm-fixed} to the ERM expert $g_\star$ with $\zeta = \frac{\lambda\rho}{64} < \frac{\lambda\tau_\D}{32}$:
\[
    \E\lrbra{\er_\D(g_\star)} \le \lr{1+\frac{\lambda}{32}}\tau_\D + Ae^{-bm}.
\]
Then apply Lemma \ref{lem:mw} with $Q$ the point mass on the expert $\star$. It is fixed in advance,
so it is determined by $S^{(0)}$ and $S^{(1)}$, and $D_{\mathrm{KL}}(Q\|\pi) = \log\frac{1}{\pi(\star)} = \log 2$.
Taking expectations and using \eqref{eq:elementary},
\[
    \E\lrbra{\er_\D(\hat h)} \le \lr{1+\frac{\lambda}{16}}\lr{1+\frac{\lambda}{32}}\tau_\D + 2Ae^{-bn/4} + \frac{128\log 2}{\lambda n}
    \le (1+\lambda)\tau_\D + K\frac{d}{\lambda n} + 2Ae^{-bn/4}.
\]
    \textbf{Conclude.} In both cases
    \[
        \E\lrbra{\er_\D(\hat h)} - (1+\lambda)\tau_\D \le K\frac{d}{\lambda n} + Ce^{-cn},
        \qquad C = 4 + 2A, \quad c = \frac{\min\lrset{\beta, b}}{4}.
    \]
    The constants $\beta$, $A$ and $b$ depend on $\D_\X$ (through $q$), on $\lambda$ and on $\H$, but not on the
    conditional distribution of $Y$ given $X$. So $C$ and $c$ are the same for all $\D$ with marginal $\D_\X$.
\end{proof}
\begin{remark}
    The result is for each fixed $\lambda>0$; the marginal-dependent remainder is not controlled uniformly as $\lambda\downarrow0$, so the theorem does not permit choosing $\lambda$ as a function of $n$ to recover an agnostic bound with $\uVCE(\mathcal H)$.
\end{remark}

\subsection{Fine-grained lower bound}
\begin{theorem}
\label{thm:relaxed-lb}
    Let $0 < \lambda \le 1$ and let $\H$ be a concept class that has an infinite uncentered $d$-VC-eluder sequence for some $d \ge 1$. There exist a marginal $\D_\X$ and sample sizes $n_1<n_2<\cdots$ such that for every learner $\A$ and every $k$ there is a distribution $\D$ with marginal $\D_\X$ satisfying
    \[
        \E_{S \sim \D^{n_k}}\lrbra{\er_\D(\A_S)}-(1+\lambda)\tau_\D \ge \frac{1}{2048}\cdot\frac{d}{\lambda n_k}.
    \]
\end{theorem}
\paragraph{Idea.} The idea is similar to the fine-grained lower bound for agnostic PAC learning.
We spread the marginal over infinitely many shattered blocks with decaying masses.
For each sample size $n_k$ we look at one block that has little probability mass, so the sample contains only a few points from it.
In this block we hide $d$ bits in noisy labels, in the spirit of Assouad's lemma \citep{assouad1983,yu1997}.
With so few points from the block, the learner cannot figure out most of the bits, and it pays for each bit it gets wrong.
The noise level is chosen proportional to $\lambda$, so that the price of the wrong bits is larger than the slack that the factor $(1+\lambda)$ gives the learner.
The earlier blocks do not help, since all the concepts we use agree there, and the later blocks have too little mass to matter.
\begin{proof}
    We first set up the three ingredients for the proof: the blocks, the witnesses, and the marginal.

    \paragraph{Setup.}
    Fix an infinite uncentered $d$-VC-eluder sequence $B_1,B_2,\ldots$ with $B_k = \{x_{k,1},\ldots, x_{k,d}\}$.
    By Definition \ref{def:vceluder}, for each $k$ there is $F_k \subseteq \H$ that VC-shatters $B_k$ and whose members all agree with each other on the earlier blocks $B_1\cup \cdots \cup B_{k-1}$.
    We record three consequences.
    \begin{itemize}
        \item For each $\theta \in \bitset^d$ there is a \emph{witness} $h_{k,\theta}\in F_k$ with $h_{k,\theta}(x_{k,j})=\theta_j$ for all $j \in [d]$.
        \item There is a function $c_k$ on $B_1\cup \cdots \cup B_{k-1}$ such that $h(x) = c_k(x)$ for every $h \in F_k$ and every $x$ in an earlier block. In words: on earlier blocks all witnesses carry the same labels, so these labels say nothing about $\theta$.
        \item The blocks are pairwise disjoint: a point of an earlier block has the same label under every member of $F_k$, so it cannot belong to the shattered set $B_k$.
    \end{itemize}
    Let $s = \frac{\lambda}{128}$, and let $\D_\X$ be the marginal that gives each point of $B_k$ mass $\frac{p_k}{d}$, where $p_k = (1-s)s^{k-1}$ is the total mass of block $k$ (note $\sum_k p_k = 1$). The total mass of the blocks after $B_k$, which we call the \emph{tail}, is
    \[
        t_k = \sum_{\ell >k}p_\ell = \frac{s}{1-s}\,p_k .
    \]
    The geometric decay is what lets one marginal serve all sample sizes: block $k$ has mass $p_k$, so it is the block that a sample of size about $\frac{1}{p_k}$ has barely explored.
    For the $k$th sample size we hide $d$ bits in block $k$.

    \paragraph{Parameters.} Take the noise level
    \[
        \gamma = \frac{\lambda}{1+2\lambda}, \qquad\text{so that}\qquad 0 < \gamma \le \frac13 \quad\text{and}\quad \frac12 - \gamma = \frac{1}{2(1+2\lambda)},
    \]
    and the sample sizes $n_k = \left\lfloor \frac{d}{128\,p_k\gamma^2} \right\rfloor$ for $k \ge 2$.
    We use three facts about these choices.
    \begin{enumerate}
        \item \emph{The tail is negligible:} since $s \le \frac12$, $t_k = \frac{s}{1-s}p_k \le 2sp_k = \frac{\lambda}{64}p_k$.
        \item \emph{$n_k$ is not too small:} for $k \ge 2$ we have $p_k \le s = \frac{\lambda}{128}$ and $\gamma \le \lambda \le 1$, so $128\,p_k\gamma^2 \le 1 \le d$, i.e., $\frac{d}{128p_k\gamma^2} \ge 1$. As $\lfloor u\rfloor \ge u/2$ for $u \ge 1$,
        \begin{equation}\label{eq:nk-size}
            \frac{d}{256\,p_k\gamma^2} \le n_k \le \frac{d}{128\,p_k\gamma^2}.
        \end{equation}
        \item \emph{The $n_k$ increase:} $p_{k+1} = sp_k \le \frac{p_k}{128}$, so by \eqref{eq:nk-size}
        $n_{k+1} \ge \frac{d}{256\,p_{k+1}\gamma^2} \ge 64\cdot\frac{d}{128\,p_k\gamma^2} \ge 64\,n_k > n_k$.
    \end{enumerate}
    Fix a learner $\A$ and $k \ge 2$.

    \paragraph{Construction.} For $\theta \in \bitset^d$ let $\D_\theta$ be the distribution with marginal $\D_\X$
    in which the label $y$ of a point $x$ is drawn as follows:
    \begin{itemize}
        \item if $x$ lies in an earlier block ($B_1,\ldots,B_{k-1}$): $y = c_k(x)$;
        \item if $x = x_{k,j}$ lies in the current block: $y = \theta_j$ with probability $\frac12 + \gamma$, and
        $y = 1 - \theta_j$ otherwise;
        \item if $x$ lies in a later block: $y = 0$.
    \end{itemize}
    Only the labels in the current block depend on $\theta$.
    This is the reason for labeling the tail by the constant $0$ rather than by $h_{k,\theta}$: the values of $h_{k,\theta}$ on later blocks are not controlled by the definition and could reveal $\theta$ to the learner.

    \paragraph{The comparator is small.} The witness $h_{k,\theta}$ never makes a mistake on earlier blocks (there it
    equals $c_k$). It makes a mistake with probability $\frac12-\gamma$ on each point of $B_k$ (there it predicts $\theta_j$),
    and may make a mistake on the tail, whose total mass is $t_k$. Since $h_{k,\theta} \in \H$, we get
    \begin{equation}\label{eq:tau-upper}
        \tau_{\D_\theta} \le \er_{\D_\theta}(h_{k,\theta}) \le p_k\lr{\frac12 - \gamma} + t_k.
    \end{equation}

    \paragraph{The learner's error on the current block.}
    Let $\xi$ denote the internal randomness of $\A$, which is independent of the sample, and let
    $q_j(\theta) = \P\lrbra{\A_S(x_{k,j}) \ne \theta_j}$ be the probability that the learner gets bit $j$
    wrong, where $S \sim \D_\theta^{n_k}$ and the probability is also over $\xi$.
    By construction, the label at $x_{k,j}$ equals $\theta_j$ with probability $\frac12+\gamma$.
    So a classifier $g:\X\to\Y$ errs at $x_{k,j}$ with probability $\frac12-\gamma$ if $g(x_{k,j}) = \theta_j$, and with probability $\frac12+\gamma$ otherwise; in both cases the error probability at $x_{k,j}$ is $\frac12 - \gamma + 2\gamma\,\1\lrbra{g(x_{k,j}) \ne \theta_j}$.
    Each point of $B_k$ has mass $\frac{p_k}{d}$, and the error on the other blocks is nonnegative, so
    \(
        \er_{\D_\theta}(g) \ge \sum_{j=1}^d \frac{p_k}{d}\lr{\frac12 - \gamma + 2\gamma\,\1\lrbra{g(x_{k,j}) \ne \theta_j}}.
    \)
    Applying this to $g = \A_S$ (which need not belong to $\H$) and taking the expectation over $S\sim\D_\theta^{n_k}$ and $\xi$,
    \begin{equation}\label{eq:block-risk}
        \E_{S\sim\D_\theta^{n_k}}\lrbra{\er_{\D_\theta}(\A_S)} \ge \sum_{j=1}^d \frac{p_k}{d}\lr{\frac12 - \gamma + 2\gamma\,q_j(\theta)}.
    \end{equation}
    It remains to show that the learner gets many bits wrong, i.e., that the $q_j(\theta)$ are large on average.

    \paragraph{The learner cannot tell $\theta_j = 0$ from $\theta_j = 1$.} Fix $j\in[d]$ and $\theta \in \bitset^d$, and let $\theta^{(j)}$ denote $\theta$ with its $j$th bit flipped.
    Let $P_\theta$ be the joint law of $(S,\xi)$ when $S \sim \D_\theta^{n_k}$, and consider the event $E = \{(S,\xi) : \A_S(x_{k,j})\neq\theta_j\}$.
    The learner is wrong about bit $j$ under $\D_\theta$ exactly when $(S,\xi) \in E$, and, because $\theta^{(j)}_j = 1-\theta_j$, it is wrong about bit $j$ under $\D_{\theta^{(j)}}$ exactly when $(S,\xi)\notin E$.
    Therefore
    \begin{equation}\label{eq:pair-sum}
        q_j(\theta) + q_j\lr{\theta^{(j)}} = P_\theta(E) + 1 - P_{\theta^{(j)}}(E) \ge 1 - \mathrm{TV}\lr{P_\theta,P_{\theta^{(j)}}} = 1 - \mathrm{TV}\lr{\D_\theta^{n_k},\D_{\theta^{(j)}}^{n_k}},
    \end{equation}
    where the last equality holds because $\xi$ has the same law under both and is independent of $S$.
    So it suffices to show that the total variation distance $\mathrm{TV}\lr{\D_\theta^{n_k},\D_{\theta^{(j)}}^{n_k}} = \sup_{E'}\lrabs{\D_\theta^{n_k}(E') - \D_{\theta^{(j)}}^{n_k}(E')}$ is small.
    We bound it through the KL divergence.

    The distributions $\D_\theta$ and $\D_{\theta^{(j)}}$ have the same marginal and differ only in the label at $x_{k,j}$: there the label equals $\theta_j$ with probability $\frac12+\gamma$ under $\D_\theta$ and with probability $\frac12-\gamma$ under $\D_{\theta^{(j)}}$.
    The labels on earlier and later blocks do not depend on $\theta$ at all.
    Since $x_{k,j}$ has mass $\frac{p_k}{d}$, the chain rule for the KL divergence gives
    \[
        \KL\lr{\D_\theta \,\|\, \D_{\theta^{(j)}}} = \frac{p_k}{d}\,\KL\lr{\mathrm{Ber}\lr{\tfrac12+\gamma}\,\|\,\mathrm{Ber}\lr{\tfrac12-\gamma}} = \frac{2\gamma\,p_k}{d}\ln\frac{1+2\gamma}{1-2\gamma} \le \frac{8\gamma^2}{1-4\gamma^2}\cdot\frac{p_k}{d} \le \frac{16\gamma^2p_k}{d},
    \]
    where we used $\ln\frac{1+u}{1-u} \le \frac{2u}{1-u^2}$ with $u = 2\gamma$, and $1-4\gamma^2 \ge \frac59 \ge \frac12$ since $\gamma \le \frac13$.
    The KL divergence is additive over the $n_k$ i.i.d.\ samples, so by the upper bound in \eqref{eq:nk-size},
    \(
        \KL\lr{\D_\theta^{n_k} \,\|\, \D_{\theta^{(j)}}^{n_k}} \le \frac{16\,n_k\,p_k\gamma^2}{d} \le \frac18 .
    \)
    By Pinsker's inequality,
    \(
        \mathrm{TV}\lr{\D_\theta^{n_k},\D_{\theta^{(j)}}^{n_k}}
        \le \sqrt{\frac12\,\KL\lr{\D_\theta^{n_k} \,\|\, \D_{\theta^{(j)}}^{n_k}}} \le \sqrt{\frac1{16}} = \frac14,
    \)
    and therefore $q_j(\theta) + q_j\lr{\theta^{(j)}} \ge \frac34$ by \eqref{eq:pair-sum}.

    \paragraph{Average over $\theta$.} The vectors $\theta \in \bitset^d$ split into $2^{d-1}$ pairs
    $\lrset{\theta,\theta^{(j)}}$, and by the previous step each pair contributes at least $\frac34$. Hence
    \[
        \frac{1}{2^d}\sum_{\theta\in\bitset^d} q_j(\theta) = \frac{1}{2^d}\sum_{\theta:\,\theta_j = 0}\lr{q_j(\theta) + q_j\lr{\theta^{(j)}}} \ge \frac{1}{2^d}\cdot 2^{d-1}\cdot\frac34 = \frac38\quad\text{for every } j \in [d].
    \]
    Averaging \eqref{eq:block-risk} over a uniformly random $\theta$ therefore gives
    \begin{equation}\label{eq:avg-risk}
        \frac{1}{2^d}\sum_\theta \E_{S\sim\D_\theta^{n_k}}\lrbra{\er_{\D_\theta}(\A_S)}
        \ge p_k\lr{\frac12 - \gamma} + 2\gamma p_k\cdot\frac38 = p_k\lr{\frac12 - \gamma} + \frac34\gamma p_k.
    \end{equation}

    \paragraph{Subtract the relaxed comparator.} Subtracting $(1+\lambda)$ times \eqref{eq:tau-upper}
    from \eqref{eq:avg-risk}, the term $p_k(\frac12-\gamma)$ cancels once and
    \[
        \frac{1}{2^d}\sum_\theta\lr{\E_{S\sim\D_\theta^{n_k}}\lrbra{\er_{\D_\theta}(\A_S)} - (1+\lambda)\tau_{\D_\theta}}
        \ge \underbrace{\frac34\gamma p_k}_{\text{wrong bits}} - \underbrace{\lambda\lr{\frac12 - \gamma}p_k}_{\text{slack on the block}}
        - \underbrace{(1+\lambda)t_k}_{\text{tail}} .
    \]
    This is where the choice of $\gamma$ is used. Since $\gamma = \frac{\lambda}{1+2\lambda}$ and $\frac12-\gamma = \frac{1}{2(1+2\lambda)}$, the wrong bits beat the slack:
    \[
        \frac34\gamma p_k - \lambda\lr{\frac12 - \gamma}p_k = \frac{3\lambda p_k}{4(1+2\lambda)} - \frac{\lambda p_k}{2(1+2\lambda)} = \frac{\lambda p_k}{4(1+2\lambda)}.
    \]
    For the tail, $1+\lambda \le 2$ and $t_k \le \frac{\lambda}{64}p_k$ give $(1+\lambda)t_k \le \frac{\lambda p_k}{32} \le \frac{\lambda p_k}{8(1+2\lambda)}$, using $1+2\lambda \le 3 \le 4$. Hence the average excess error is at least
    \[
        \frac{\lambda p_k}{4(1+2\lambda)} - \frac{\lambda p_k}{8(1+2\lambda)} = \frac{\lambda p_k}{8(1+2\lambda)}.
    \]
    Finally, the lower bound in \eqref{eq:nk-size} gives $p_k \ge \frac{d}{256\gamma^2n_k} = \frac{d(1+2\lambda)^2}{256\lambda^2n_k}$, so
    \[
        \frac{\lambda p_k}{8(1+2\lambda)} \ge \frac{d\,(1+2\lambda)}{2048\,\lambda n_k} \ge \frac{d}{2048\,\lambda n_k}.
    \]
    Some $\theta$ attains at least the average; take $\D = \D_\theta$.

    The sample sizes $n_k$ are defined and increasing for $k \ge 2$. Relabeling them as $n_1 < n_2 < \cdots$ gives the statement of the theorem.
\end{proof}

\section{Conclusion}

We studied marginal-nonuniform learning in the agnostic case, for binary classification, at the level of both rates and fine-grained rates. 
In the standard agnostic setting, allowing the constants to depend on the marginal does not help: a class is learnable at rate $\frac{1}{\sqrt n}$ exactly when its VC dimension is finite, and the fine-grained rate is $\sqrt{\VC(\H)/n}$, as in uniform PAC learning, with a lower bound attained by a single marginal. 
In the relaxed setting, where the learner competes with $(1+\lambda)\inf_{h\in\H}\er_\D(h)$, the picture of the realizable case returns. 
The rates form the same trichotomy as in the realizable case: exponential for finite classes, linear for infinite classes of finite VC dimension, and arbitrarily slow otherwise. 
The fine-grained rate is of order $\frac{d}{\lambda n}$, where $d$ is the uncentered VC-eluder dimension, and the marginal only affects an exponentially small term. 
So the slow rate of the agnostic setting comes only from competing with the best concept exactly: any constant multiplicative slack restores the realizable rates, and even makes finite classes learnable at an exponential rate.

We expect the ideas of this paper to extend to multiclass classification with a bounded label space. 
The lower bounds only use the labels of a shattered set, and the upper bound is built from three ingredients that are not specific to two labels: the exhaustion lemma, which \cite{brukhim2026marginal} already prove in the multiclass setting; the version spaces of clean subsamples; and aggregation by exponential weights. 
With the VC dimension replaced by a suitable multiclass dimension and a multiclass first-order learner in place of the binary one, we believe the same arguments go through, with constants that depend on the number of labels. The main open problem is the case of an infinite label space, where two dimensions play a role \citep{cohen2026agnostic}. 

\bibliography{bibliography}

\begin{thebibliography}{40}
\providecommand{\natexlab}[1]{#1}
\providecommand{\url}[1]{\texttt{#1}}
\expandafter\ifx\csname urlstyle\endcsname\relax
  \providecommand{\doi}[1]{doi: #1}\else
  \providecommand{\doi}{doi: \begingroup \urlstyle{rm}\Url}\fi

\bibitem[Asilis et~al.(2025)Asilis, H{\o}gsgaard, and Velegkas]{asilis2025agnostic}
Julian Asilis, Mikael~M{\o}ller H{\o}gsgaard, and Grigoris Velegkas.
\newblock On agnostic {PAC} learning in the small error regime.
\newblock In \emph{Advances in Neural Information Processing Systems}, volume~38, 2025.

\bibitem[Assouad(1983)]{assouad1983}
Patrice Assouad.
\newblock Deux remarques sur l'estimation.
\newblock \emph{Comptes Rendus de l'Acad{\'e}mie des Sciences de Paris, S{\'e}rie I}, 296:\penalty0 1021--1024, 1983.

\bibitem[Awasthi et~al.(2017)Awasthi, Balcan, and Long]{awasthi2018powerlocalizationefficientlylearning}
Pranjal Awasthi, Maria~Florina Balcan, and Philip~M. Long.
\newblock The power of localization for efficiently learning linear separators with noise.
\newblock \emph{Journal of the ACM}, 63\penalty0 (6):\penalty0 50:1--50:27, 2017.

\bibitem[Ben-David et~al.(1995)Ben-David, Benedek, and Mansour]{bendavid1995}
Shai Ben-David, Gyora~M. Benedek, and Yishay Mansour.
\newblock A parameterization scheme for classifying models of {PAC} learnability.
\newblock \emph{Inf. Comput.}, 120\penalty0 (1):\penalty0 11–21, July 1995.
\newblock ISSN 0890-5401.
\newblock \doi{10.1006/inco.1995.1094}.
\newblock URL \url{https://doi.org/10.1006/inco.1995.1094}.

\bibitem[Ben-David et~al.(2008)Ben-David, Lu, and P{\'a}l]{bendavid2008does}
Shai Ben-David, Tyler Lu, and D{\'a}vid P{\'a}l.
\newblock Does unlabeled data provably help? {W}orst-case analysis of the sample complexity of semi-supervised learning.
\newblock In \emph{Proceedings of the 21st Annual Conference on Learning Theory (COLT)}, pages 33--44, 2008.

\bibitem[Benedek and Itai(1991)]{benedek1991learnability}
Gyora~M. Benedek and Alon Itai.
\newblock Learnability with respect to fixed distributions.
\newblock \emph{Theoretical Computer Science}, 86\penalty0 (2):\penalty0 377--389, 1991.

\bibitem[Benedek and Itai(1994)]{benedek1994}
Gyora~M. Benedek and Alon Itai.
\newblock Nonuniform learnability.
\newblock \emph{Journal of Computer and System Sciences}, 48\penalty0 (2):\penalty0 311--323, 1994.

\bibitem[Blumer et~al.(1987)Blumer, Ehrenfeucht, Haussler, and Warmuth]{blumer1987}
Anselm Blumer, Andrzej Ehrenfeucht, David Haussler, and Manfred~K. Warmuth.
\newblock {O}ccam's razor.
\newblock \emph{Information Processing Letters}, 24\penalty0 (6):\penalty0 377--380, 1987.

\bibitem[Bousquet et~al.(2021)Bousquet, Hanneke, Moran, van Handel, and Yehudayoff]{bousquet2020theoryuniversallearning}
Olivier Bousquet, Steve Hanneke, Shay Moran, Ramon van Handel, and Amir Yehudayoff.
\newblock A theory of universal learning.
\newblock In \emph{Proceedings of the 53rd Annual ACM SIGACT Symposium on Theory of Computing (STOC)}, pages 532--541, 2021.

\bibitem[Bousquet et~al.(2023)Bousquet, Hanneke, Moran, Shafer, and Tolstikhin]{bousquet2022finegraineddistributiondependentlearningcurves}
Olivier Bousquet, Steve Hanneke, Shay Moran, Jonathan Shafer, and Ilya Tolstikhin.
\newblock Fine-grained distribution-dependent learning curves.
\newblock In \emph{Proceedings of the 36th Conference on Learning Theory}, volume 195 of \emph{Proceedings of Machine Learning Research}, pages 5890--5924, 2023.

\bibitem[Brukhim et~al.(2026)Brukhim, Hanneke, Shaeiri, and Thiessen]{brukhim2026marginal}
Nataly Brukhim, Steve Hanneke, Amirreza Shaeiri, and Maximilian Thiessen.
\newblock Marginal-nonuniform multiclass learning.
\newblock In \emph{Advances in Neural Information Processing Systems}, 2026.

\bibitem[Cesa-Bianchi et~al.(2004)Cesa-Bianchi, Conconi, and Gentile]{cesabianchi2004generalization}
Nicol{\`o} Cesa-Bianchi, Alex Conconi, and Claudio Gentile.
\newblock On the generalization ability of on-line learning algorithms.
\newblock \emph{IEEE Transactions on Information Theory}, 50\penalty0 (9):\penalty0 2050--2057, 2004.

\bibitem[Cohen et~al.(2026)Cohen, Erez, Hanneke, Koren, Mansour, Moran, and Zhang]{cohen2026agnostic}
Alon Cohen, Liad Erez, Steve Hanneke, Tomer Koren, Yishay Mansour, Shay Moran, and Qian Zhang.
\newblock Sample complexity of agnostic multiclass classification: {N}atarajan dimension strikes back.
\newblock In \emph{Proceedings of the 58th Annual ACM Symposium on Theory of Computing}, STOC '26, page 722–733, New York, NY, USA, 2026. Association for Computing Machinery.
\newblock ISBN 9798400725364.
\newblock \doi{10.1145/3798129.3800787}.
\newblock URL \url{https://doi.org/10.1145/3798129.3800787}.

\bibitem[Daniely(2015)]{daniely2015}
Amit Daniely.
\newblock A {PTAS} for agnostically learning halfspaces.
\newblock In \emph{Proceedings of The 28th Conference on Learning Theory}, volume~40 of \emph{Proceedings of Machine Learning Research}, pages 484--502. PMLR, 2015.

\bibitem[Darnst{\"a}dt et~al.(2013)Darnst{\"a}dt, Simon, and Sz{\"o}r{\'e}nyi]{darnstadt2013unlabeled}
Malte Darnst{\"a}dt, Hans~Ulrich Simon, and Bal{\'a}zs Sz{\"o}r{\'e}nyi.
\newblock Unlabeled data does provably help.
\newblock In \emph{30th International Symposium on Theoretical Aspects of Computer Science (STACS)}, volume~20 of \emph{Leibniz International Proceedings in Informatics (LIPIcs)}, pages 185--196, 2013.

\bibitem[David et~al.(2016)David, Moran, and Yehudayoff]{david2016supervised}
Ofir David, Shay Moran, and Amir Yehudayoff.
\newblock Supervised learning through the lens of compression.
\newblock In \emph{Advances in Neural Information Processing Systems}, volume~29, 2016.

\bibitem[Devroye et~al.(1996)Devroye, Gy{\"o}rfi, and Lugosi]{devroye1996}
Luc Devroye, L{\'a}szl{\'o} Gy{\"o}rfi, and G{\'a}bor Lugosi.
\newblock \emph{A Probabilistic Theory of Pattern Recognition}, volume~31 of \emph{Applications of Mathematics}.
\newblock Springer, 1996.

\bibitem[Freund and Schapire(1997)]{freund1997decision}
Yoav Freund and Robert~E. Schapire.
\newblock A decision-theoretic generalization of on-line learning and an application to boosting.
\newblock \emph{Journal of Computer and System Sciences}, 55\penalty0 (1):\penalty0 119--139, 1997.

\bibitem[G{\"o}pfert et~al.(2019)G{\"o}pfert, Ben-David, Bousquet, Gelly, Tolstikhin, and Urner]{gopfert2019when}
Christina G{\"o}pfert, Shai Ben-David, Olivier Bousquet, Sylvain Gelly, Ilya Tolstikhin, and Ruth Urner.
\newblock When can unlabeled data improve the learning rate?
\newblock In \emph{Proceedings of the Thirty-Second Conference on Learning Theory}, volume~99 of \emph{Proceedings of Machine Learning Research}, pages 1500--1518. PMLR, 2019.

\bibitem[Hanneke and Moran(2026)]{hanneke2026universalagnostic}
Steve Hanneke and Shay Moran.
\newblock A theory of universal agnostic learning.
\newblock \emph{arXiv preprint arXiv:2601.20961}, 2026.

\bibitem[Hanneke and Xu(2024)]{hanneke2025universalratesempiricalrisk}
Steve Hanneke and Mingyue Xu.
\newblock Universal rates of empirical risk minimization.
\newblock In \emph{Advances in Neural Information Processing Systems}, volume~37, 2024.

\bibitem[Hanneke and Xu(2025)]{hanneke2025ermagnostic}
Steve Hanneke and Mingyue Xu.
\newblock Universal rates of {ERM} for agnostic learning.
\newblock In \emph{Proceedings of the Thirty Eighth Conference on Learning Theory}, volume 291 of \emph{Proceedings of Machine Learning Research}, pages 2666--2703. PMLR, 2025.

\bibitem[Hanneke et~al.(2024)Hanneke, Larsen, and Zhivotovskiy]{hanneke2024revisitingagnosticpaclearning}
Steve Hanneke, Kasper~Green Larsen, and Nikita Zhivotovskiy.
\newblock Revisiting agnostic {PAC} learning.
\newblock In \emph{Proceedings of the 65th IEEE Annual Symposium on Foundations of Computer Science (FOCS)}, 2024.

\bibitem[Hanneke et~al.(2025{\natexlab{a}})Hanneke, Meng, and Shaeiri]{hanneke2025representation}
Steve Hanneke, Qinglin Meng, and Amirreza Shaeiri.
\newblock Representation preserving multiclass agnostic to realizable reduction.
\newblock In \emph{ICML}, 2025{\natexlab{a}}.

\bibitem[Hanneke et~al.(2025{\natexlab{b}})Hanneke, Moran, and Thiessen]{hanneke2025marginalnonuniform}
Steve Hanneke, Shay Moran, and Maximilian Thiessen.
\newblock Marginal-nonuniform {PAC} learnability.
\newblock In \emph{The Thirty-ninth Annual Conference on Neural Information Processing Systems}, 2025{\natexlab{b}}.
\newblock URL \url{https://openreview.net/forum?id=aoVCFtox89}.

\bibitem[Hanneke et~al.(2025{\natexlab{c}})Hanneke, Shaeiri, and Wang]{hanneke2025nonuniformbandit}
Steve Hanneke, Amirreza Shaeiri, and Hongao Wang.
\newblock Non-uniform multiclass learning with bandit feedback.
\newblock In \emph{Advances in Neural Information Processing Systems}, volume~38, 2025{\natexlab{c}}.

\bibitem[Haussler(1992)]{Haussler1992}
David Haussler.
\newblock Decision theoretic generalizations of the {PAC} model for neural net and other learning applications.
\newblock \emph{Information and Computation}, 100\penalty0 (1):\penalty0 78–150, 1992.
\newblock ISSN 0890-5401.
\newblock \doi{10.1016/0890-5401(92)90010-d}.
\newblock URL \url{http://dx.doi.org/10.1016/0890-5401(92)90010-D}.

\bibitem[Hopkins et~al.(2022)Hopkins, Kane, Lovett, and Mahajan]{hopkins2022realizable}
Max Hopkins, Daniel~M. Kane, Shachar Lovett, and Gaurav Mahajan.
\newblock Realizable learning is all you need.
\newblock In \emph{Proceedings of the 35th Conference on Learning Theory}, volume 178 of \emph{Proceedings of Machine Learning Research}, 2022.

\bibitem[Kearns et~al.(1994)Kearns, Schapire, and Sellie]{kearns1994}
Michael~J. Kearns, Robert~E. Schapire, and Linda~M. Sellie.
\newblock Toward efficient agnostic learning.
\newblock \emph{Machine Learning}, 17\penalty0 (2--3):\penalty0 115--141, 1994.

\bibitem[Lecu{\'e} and Mendelson(2012)]{lecue2012general}
Guillaume Lecu{\'e} and Shahar Mendelson.
\newblock General nonexact oracle inequalities for classes with a subexponential envelope.
\newblock \emph{The Annals of Statistics}, 40\penalty0 (2):\penalty0 832--860, 2012.

\bibitem[Littlestone and Warmuth(1994)]{littlestone1994}
Nick Littlestone and Manfred~K. Warmuth.
\newblock The weighted majority algorithm.
\newblock \emph{Information and Computation}, 108\penalty0 (2):\penalty0 212--261, 1994.

\bibitem[Lugosi and Zeger(1996)]{lugosi1996}
G{\'a}bor Lugosi and Kenneth Zeger.
\newblock Concept learning using complexity regularization.
\newblock \emph{IEEE Transactions on Information Theory}, 42\penalty0 (1):\penalty0 48--54, 1996.

\bibitem[Mathiasen et~al.(2026)Mathiasen, Qian, and Zhivotovskiy]{mathiasen2026optimalagnosticpacalgorithm}
Markus~Engelund Mathiasen, Jian Qian, and Nikita Zhivotovskiy.
\newblock An optimal agnostic {PAC} algorithm, 2026.
\newblock URL \url{https://arxiv.org/abs/2608.06363}.

\bibitem[Rissanen(1978)]{rissanen1978}
Jorma Rissanen.
\newblock Modeling by shortest data description.
\newblock \emph{Automatica}, 14\penalty0 (5):\penalty0 465--471, 1978.

\bibitem[Shalev-Shwartz and Ben-David(2014)]{shalev2014understanding}
Shai Shalev-Shwartz and Shai Ben-David.
\newblock \emph{Understanding Machine Learning: From Theory to Algorithms}.
\newblock Cambridge University Press, New York, NY, USA, 2014.
\newblock ISBN 978-1-107-05713-5.

\bibitem[Srebro et~al.(2010)Srebro, Sridharan, and Tewari]{srebro2010}
Nathan Srebro, Karthik Sridharan, and Ambuj Tewari.
\newblock Smoothness, low noise and fast rates.
\newblock In \emph{Advances in Neural Information Processing Systems}, volume~23, pages 2199--2207, 2010.

\bibitem[Valiant(1984)]{valiant1984}
L.~G. Valiant.
\newblock A theory of the learnable.
\newblock \emph{Commun. ACM}, 27\penalty0 (11):\penalty0 1134–1142, November 1984.
\newblock ISSN 0001-0782.
\newblock \doi{10.1145/1968.1972}.
\newblock URL \url{https://doi.org/10.1145/1968.1972}.

\bibitem[Vapnik and Chervonenkis(1971)]{vc1971}
V.~N. Vapnik and A.~Ya. Chervonenkis.
\newblock On the uniform convergence of relative frequencies of events to their probabilities.
\newblock \emph{Theory of Probability \& Its Applications}, 16\penalty0 (2):\penalty0 264--280, 1971.
\newblock \doi{10.1137/1116025}.
\newblock URL \url{https://doi.org/10.1137/1116025}.

\bibitem[Vapnik(1998)]{vapnik1998}
Vladimir~N. Vapnik.
\newblock \emph{Statistical Learning Theory}.
\newblock Wiley-Interscience, 1998.

\bibitem[Yu(1997)]{yu1997}
Bin Yu.
\newblock {A}ssouad, {F}ano, and {L}e {C}am.
\newblock In \emph{Festschrift for Lucien Le Cam}, pages 423--435. Springer, 1997.

\end{thebibliography}

\appendix

\section{Existence of the first-order learner}
\label{appendix:firstorderlearner}
Our upper bounds run a learner inside a version space of $\H$, and they need a guarantee whose lower-order term scales with the complexity of that version space. 
Known first-order guarantees for agnostic PAC learning \citep{hanneke2024revisitingagnosticpaclearning,mathiasen2026optimalagnosticpacalgorithm} assume that the class has VC dimension at most $k$. 
This is not what we have. 
The version spaces used in section \ref{sec:upper} are only simple in the distributional sense, $\VC_{\D_\X}(\V) \le k$, and their VC dimension can be much larger than $k$. 
For instance, let $\X = \{z_1,\ldots,z_M\} \cup \N$, and let $\H$ consist of all labelings of $z_1,\ldots,z_M$ that are $0$ on $\N$, together with the indicator functions $\1_{\{x\}}$ of the points $x \in \N$. 
Then $\VC(\H) = M$ while $\uVCE(\H) = 1$. 
If $\D_\X$ puts no mass on the points $z_j$, then no sample contains them, so every version space still VC-shatters $\{z_1,\ldots,z_M\}$. 
Every version space thus has VC dimension $M$, while its distributional VC dimension is at most $1$. 
Applying a first-order guarantee with $k = M$ would give a bound of order $\frac{\VC(\H)}{\lambda n}$ instead of $\frac{\uVCE(\H)}{\lambda n}$. 
Moreover, the learner does not know $\D_\X$, so it cannot discard the part of the class that lives on a set of probability zero. 
We therefore need a first-order learner whose guarantee depends only on the distributional VC dimension. 
Definition~\ref{def:firstorderlearner} asks for such a learner, and Lemma~\ref{lem:first-order-exists} shows that it exists.
The point is that a shattered set of probability zero almost surely never appears among the sample points or the test point, so a learner whose guarantee depends only on the points it sees is not affected by it.
\begin{lemma}\label{lem:first-order-exists}
A first-order learner in the sense of Definition \ref{def:firstorderlearner} exists.
\end{lemma}

\begin{proof}
We use a rule $\mathcal B$ with the following leave-one-out guarantee \citep{mathiasen2026optimalagnosticpacalgorithm}: there are universal constants $c_1, c_2$ such that for every $N$, every labeled sequence $z_1,\ldots,z_N$ with $z_i = (x_i,y_i)$, and every class $\V$ whose trace on $(x_1,\ldots,x_N)$ has VC dimension at most $k$,
\[
    \frac1N\sum_{i=1}^N \P\lrbra{\mathcal B_\V(z_{-i})(x_i) \ne y_i} \le L^* + c_1\sqrt{\frac{L^*k}{N}} + c_2\frac{k}{N},
    \qquad L^* = \min_{h\in\V}\frac1N\sum_{i=1}^N \1[h(x_i)\ne y_i],
\]
where $z_{-i}$ is the sequence with $z_i$ removed and the probability is over the randomness of $\mathcal B$.

Now let $S \sim \D^n$ and draw a fresh $(X,Y)\sim\D$, so that $z_1,\ldots,z_{n+1}$ are i.i.d.\ from $\D$.

\emph{The trace condition holds almost surely.} If $\VC_{\D_\X}(\V)\le k$, then each of the
$\binom{n+1}{k+1}$ sub-tuples of size $k+1$ of $(x_1,\ldots,x_{n+1})$ is shattered by $\V$ with probability
$0$, so the trace of $\V$ on these points has VC dimension at most $k$ almost surely.

\emph{Exchangeability.} Since the $z_i$ are i.i.d., each term of the leave-one-out average has the same
expectation as the error of $\mathcal B_\V(S)$ on the fresh point, so
$\E\lrbra{\er_\D(\mathcal B_\V(S))}$ equals the expected left-hand side above with $N = n+1$.

\emph{Jensen.} Taking expectations and using concavity of the square root,
\[
    \E\lrbra{\er_\D(\mathcal B_\V(S))} \le \E[L^*] + c_1\sqrt{\frac{\E[L^*]\,k}{n+1}} + c_2\frac{k}{n+1}.
\]
Finally, for every $h\in\V$ we have $\E[L^*] \le \er_\D(h)$, hence $\E[L^*] \le \tau_{\D,\V}$. This gives
\eqref{eq:first-order} with $B = \max\{c_1,c_2,1\}$.
\end{proof}
\begin{remark}
\citet{hanneke2024revisitingagnosticpaclearning} were the first to obtain a first-order bound with the
optimal $\sqrt{\tau_{\D,\V}\,k/n}$ term, with an additional polylogarithmic factor in the lower-order
term. Using their learner in place of $\mathcal B$, the same argument gives
$\E\lrbra{\er_\D(\A_{\lambda,S})} - (1+\lambda)\tau_\D \le K\,\frac{d\,\mathrm{polylog}(n/d)}{\lambda n} + Ce^{-cn}$.
The optimal learner of \citet{mathiasen2026optimalagnosticpacalgorithm} removes the logarithmic factor,
which is what gives the exact fine-grained rate $\frac{d}{\lambda n}$ and the linear rate in
Theorem \ref{thm:relaxed-rates}.
\end{remark}
\section{Measurability}
\label{appendix:measurability}
As in \citet{bousquet2020theoryuniversallearning} and
\citet{hanneke2025marginalnonuniform}, we assume that $\X$ is a Polish space with its Borel
$\sigma$-algebra and that $\H$ is measurable in the following sense: there are a Polish space $\Theta$
and a Borel map $(\theta,x)\mapsto h_\theta(x)$ such that $\H = \{h_\theta : \theta\in\Theta\}$. Under
this assumption, every event in this paper that is defined by an existential quantifier over $\H$ (for
example, that a tuple of points is VC-shattered by a version space) is an analytic set, hence universally
measurable, so all the probabilities we write are well defined. We also assume that the learning rules we
use, namely ERM and the first-order learner, are universally measurable functions of the data; for ERM
such a choice exists by the Jankov--von Neumann uniformization theorem. If $\H$ is countable, all of this
holds trivially.
\section{The uncentered VC-eluder dimension}\label{appendix:centeredvsuncentered}

In this appendix we collect the properties of $\uVCE$ that we use, and we show that for binary
classification it coincides with the VC-eluder dimension of \citet{hanneke2025universalratesempiricalrisk},
which characterizes the realizable marginal-nonuniform rate \citep{hanneke2025marginalnonuniform}.
All statements here are purely combinatorial.

We first recall the centered notion. Following \citet{hanneke2025universalratesempiricalrisk}, an
\emph{infinite $d$-VC-eluder sequence} for $\H$ is a sequence of labeled points
$(x_1,y_1),(x_2,y_2),\ldots$ such that every finite prefix is realizable by $\H$ and, for every $k\ge1$,
the block $\{x_{(k-1)d+1},\ldots,x_{kd}\}$ is VC-shattered by the version space
$\lrset{h\in\H : h(x_i) = y_i \text{ for all } i \le (k-1)d}$. The \emph{VC-eluder dimension} $\VCE(\H)$ is
the largest $d$ for which such a sequence exists, with $\VCE(\H) = 0$ if there is none for $d = 1$ and
$\VCE(\H)=\infty$ if there is one for every $d$.

The two notions differ in what the witnesses must agree on. In the centered version, all concepts that
shatter a block agree with one fixed labeling of the earlier blocks. In the uncentered version
(Definition~\ref{def:vceluder}), the concepts that shatter block $k$ only need to agree with each other on
the earlier blocks, and different blocks may use different labelings of the past. For binary labels this
extra freedom does not help.

\begin{proposition}\label{prop:uvce-vce}
For every concept class $\H \subseteq \{0,1\}^\X$ and every $d \ge 1$, $\H$ has an infinite uncentered
$d$-VC-eluder sequence if and only if it has an infinite $d$-VC-eluder sequence. In particular,
$\uVCE(\H) = \VCE(\H)$.
\end{proposition}

\begin{proof}
\emph{Centered implies uncentered.} Given an infinite $d$-VC-eluder sequence, let $B_k$ be its $k$th block
and $\mathcal F_k$ the version space of the labeled points before $B_k$. Then $\mathcal F_k$ shatters $B_k$,
and all its members agree with the labels $y_i$ on the earlier blocks, so in particular they agree with each
other there. Hence $B_1, B_2, \ldots$ is an infinite uncentered $d$-VC-eluder sequence.

\emph{Uncentered implies centered.} Let $B_1, B_2,\ldots$ be an infinite uncentered $d$-VC-eluder sequence.
For each $k$ let $\mathcal F_k \subseteq \H$ shatter $B_k$ with all members agreeing with each other on
$B_1\cup\cdots\cup B_{k-1}$, and let $c_k$ be their common labeling there. We pass to a subsequence along
which these labelings are consistent. Let $N_0 = \N$. Given an infinite set $N_{j-1}\subseteq\N$, let
$k_j = \min N_{j-1}$. For every $k \in N_{j-1}$ with $k > k_j$, the restriction of $c_k$ to $B_{k_j}$ is one
of the $2^d$ labelings of $B_{k_j}$, so some labeling $\sigma_j$ occurs for infinitely many such $k$. Let
$N_j$ be the set of these $k$; it is infinite, and $N_j \subseteq N_{j-1}$. Since $k_j \in N_{j-1}\subseteq N_i$
for all $i<j$, we have
\begin{equation}\label{eq:consistent-labels}
    c_{k_j} = \sigma_i \ \text{on } B_{k_i} \qquad \text{for all } i < j.
\end{equation}
Now list the blocks $B_{k_1}, B_{k_2},\ldots$ in this order, labeling the points of $B_{k_i}$ by $\sigma_i$.
We check that this is an infinite $d$-VC-eluder sequence.
\begin{itemize}
    \item \emph{Every block is shattered by the current version space.} Let $\V_j$ be the set of concepts
    that equal $\sigma_i$ on $B_{k_i}$ for all $i<j$. By \eqref{eq:consistent-labels}, every member of
    $\mathcal F_{k_j}$ lies in $\V_j$, and $\mathcal F_{k_j}$ shatters $B_{k_j}$. Hence $\V_j$ shatters $B_{k_j}$.
    \item \emph{Every finite prefix is realizable.} Any member of $\mathcal F_{k_{j+1}}$ equals $\sigma_i$ on
    $B_{k_i}$ for all $i \le j$, by \eqref{eq:consistent-labels} with $j+1$ in place of $j$. So the labeled
    blocks $B_{k_1},\ldots,B_{k_j}$, and hence every prefix of the sequence, are realizable by $\H$.
\end{itemize}
Both notions are monotone in $d$ (dropping one point from every block gives a sequence of width $d-1$), so
the equivalence for every $d$ gives $\uVCE(\H) = \VCE(\H)$, including the cases $0$ and $\infty$.
\end{proof}

The only place where binary labels enter is the pigeonhole step: a block of size $d$ has finitely many
labelings. The same argument works for any finite label space, with $|\Y|^d$ in place of $2^d$, but not for
an infinite label space. This is why we state our results with the uncentered notion, which is also the form
used in the multiclass setting \citep{brukhim2026marginal}, and which is the form our proofs use directly
(Lemma~\ref{lem:exhaustion} and Theorem~\ref{thm:relaxed-lb}).

\begin{remark}\label{rem:same-as-realizable}
By Proposition~\ref{prop:uvce-vce}, the fine-grained rate $\frac{\uVCE(\H)}{\lambda n}$ of
Theorem~\ref{thm:relaxed} equals $\frac{\VCE(\H)}{\lambda n}$. The realizable fine-grained rate is
$\frac{\VCE(\H)}{n}$ \citep{hanneke2025marginalnonuniform}. So under the relaxed criterion, the agnostic
fine-grained rate is governed by the same dimension as the realizable one, and only an additional factor
$\frac1\lambda$ is paid.
\end{remark}

We next relate $\uVCE$ to the size and the VC dimension of the class.

\begin{lemma}\label{lem:uvce-basic2}
For every concept class $\H$:
\begin{enumerate}
    \item $\uVCE(\H) \le \VC(\H)$;
    \item $\uVCE(\H) = 0$ if and only if $\H$ is finite;
    \item $\uVCE(\H) < \infty$ if and only if $\VC(\H) < \infty$.
\end{enumerate}
\end{lemma}

\begin{proof}
\emph{1.} The first block of an infinite uncentered $d$-VC-eluder sequence is a VC-shattered set of size
$d$.

\emph{2, finite classes.} Suppose $\H$ has an infinite uncentered $1$-VC-eluder sequence $B_1,B_2,\ldots$
with witnesses $\mathcal F_k$. Since $\mathcal F_k$ shatters $B_k$, it contains two concepts $f_k \ne g_k$
that disagree on $B_k$. For $k < k'$, the concepts $f_{k'}$ and $g_{k'}$ agree on $B_k$, while $f_k$ and
$g_k$ do not, so $\{f_k,g_k\} \ne \{f_{k'},g_{k'}\}$. Hence $\H$ contains infinitely many distinct pairs of
concepts, and $\H$ is infinite. So a finite class has $\uVCE(\H) = 0$.

\emph{2, infinite classes.} We build an infinite uncentered $1$-VC-eluder sequence. Let $\H_0 = \H$. Given
an infinite class $\H_{r-1}\subseteq\H$ whose members agree with each other on the points
$x_1,\ldots,x_{r-1}$ chosen so far, pick two members of $\H_{r-1}$ and a point $x_r$ where they disagree.
Then $\{x_r\}$ is VC-shattered by $\H_{r-1}$, and $x_r$ is a new point, since all members of $\H_{r-1}$
agree on the earlier points. One of the two classes $\{h \in \H_{r-1} : h(x_r) = 0\}$ and
$\{h \in \H_{r-1} : h(x_r) = 1\}$ is infinite; call it $\H_r$. Its members agree on $x_1,\ldots,x_r$, so we
can continue. The singletons $\{x_1\},\{x_2\},\ldots$ form an infinite uncentered $1$-VC-eluder sequence.

\emph{3.} If $\VC(\H) < \infty$, then $\uVCE(\H) < \infty$ by item 1. Conversely, suppose
$\VC(\H) = \infty$ and fix $d \ge 1$; we build an infinite uncentered $d$-VC-eluder sequence. The argument is
the same as for item 2, with ``infinite VC dimension'' in place of ``infinite''. Let $\H_0 = \H$. Given
$\H_{r-1}\subseteq\H$ with $\VC(\H_{r-1}) = \infty$ whose members agree with each other on
$B_1\cup\cdots\cup B_{r-1}$, pick a set $B_r$ of size $d$ that is VC-shattered by $\H_{r-1}$. It is disjoint
from the earlier blocks, since all members of $\H_{r-1}$ agree there. Split $\H_{r-1}$ into the $2^d$ classes
of concepts with a given labeling of $B_r$. Since the VC dimension of a union of two classes is at most the
sum of their VC dimensions plus one, a finite union of classes of finite VC dimension has finite VC
dimension. So one of these $2^d$ classes has infinite VC dimension; call it $\H_r$. Its members agree on
$B_1\cup\cdots\cup B_r$, so we can continue. Hence $\uVCE(\H) \ge d$ for every $d$, i.e.,
$\uVCE(\H) = \infty$.
\end{proof}

By Proposition~\ref{prop:uvce-vce}, items 2 and 3 are also consequences of the corresponding facts for
$\VCE$ \citep{hanneke2025universalratesempiricalrisk}; we include the short proofs to keep the paper
self-contained.

Finally, $\uVCE(\H)$ can be much smaller than $\VC(\H)$, which is where the relaxed criterion gains over
the standard one.

\begin{example}\label{ex:uvce-small}
Let $M \ge 2$, let $\X = \{z_1,\ldots,z_M\} \cup \N$, and let $\H$ consist of
\begin{itemize}
    \item all labelings of $z_1,\ldots,z_M$ that are $0$ on $\N$, and
    \item the indicator functions $\1_{\{x\}}$ of the points $x\in\N$.
\end{itemize}
The set $\{z_1,\ldots,z_M\}$ is VC-shattered. No set that contains a point of $\N$ together with another
point is VC-shattered, since no concept labels both points $1$. Hence $\VC(\H) = M$. The class is infinite,
so $\uVCE(\H) \ge 1$ by Lemma~\ref{lem:uvce-basic2}. In an infinite uncentered $2$-VC-eluder sequence,
every block would be a shattered pair, hence a subset of $\{z_1,\ldots,z_M\}$, and the blocks are pairwise
disjoint, so there are at most $M/2$ of them. Hence $\uVCE(\H) = 1$. For this class, the standard agnostic
fine-grained rate is $\sqrt{M/n}$ (Theorem~\ref{thm:standard-fine}), while the relaxed one is
$\frac{1}{\lambda n}$ (Theorem~\ref{thm:relaxed}), whatever $M$ is. The finite part of the class only
affects the marginal-dependent lower-order term.
\end{example}
\end{document}